\documentclass[10pt,twocolumn,letterpaper]{article}

\usepackage{cvpr}
\usepackage{times}
\usepackage{epsfig}
\usepackage{graphicx}
\usepackage{amsmath}
\usepackage{amssymb}
\usepackage{xspace}
\usepackage{booktabs}
\usepackage[ruled,vlined]{algorithm2e}
\usepackage{multirow}
\usepackage{xcolor}
\usepackage{authblk}

\usepackage[pagebackref=true,breaklinks=true,letterpaper=true,colorlinks,bookmarks=false]{hyperref}

\cvprfinalcopy 

\newcommand{\algLabel}{Algorithm\xspace}
\newcommand{\figLabel}{Figure\xspace}
\newcommand{\eqLabel}{Equation\xspace}
\newcommand{\secLabel}{Section\xspace}
\newcommand{\tblLabel}{Table\xspace}
\newcommand{\mysection}[1]{\vspace{3pt}\noindent\textbf{#1.}}
\newcommand{\supp}{\textbf{supplement material}\xspace}
\newcommand{\secor}{degenerate search-evaluation correlation}

\newcommand{\savespace}{\vspace{0pt}}

\begin{document}

\renewcommand\Authsep{, }
\renewcommand\Authand{}
\renewcommand\Authands{, }
\newcommand\CoAuthorMark{\footnotemark[\arabic{footnote}]}
\title{SGAS: Sequential Greedy Architecture Search  \\ \small\url{https://www.deepgcns.org/auto/sgas} \vspace{-20pt}}

\author[1]{Guohao Li\thanks{equal contribution}\vspace{-10pt}}
\author[1]{Guocheng Qian\protect\CoAuthorMark}
\author[1]{Itzel C. Delgadillo\protect\CoAuthorMark}
\author[2]{Matthias M\"uller}
\author[1]{Ali Thabet}
\author[1]{Bernard Ghanem}
\affil[1]{King Abdullah University of Science and Technology (KAUST), Saudi Arabia}
\affil[2]{Intelligent Systems Lab,~ Intel Labs,~ Germany}

\maketitle

\begin{abstract}
Architecture design has become a crucial component of successful deep learning. Recent progress in automatic neural architecture search (NAS) shows a lot of promise. However, discovered architectures often fail to generalize in the final evaluation. Architectures with a higher validation accuracy during the search phase may perform worse in the evaluation (see \figLabel \ref{fig:intro_fig}). Aiming to alleviate this common issue, we introduce sequential greedy architecture search (SGAS), an efficient method for neural architecture search. By dividing the search procedure into sub-problems, SGAS chooses and prunes candidate operations in a greedy fashion. We apply SGAS to search architectures for Convolutional Neural Networks (CNN) and Graph Convolutional Networks (GCN). Extensive experiments show that SGAS is able to find state-of-the-art architectures for tasks such as image classification, point cloud classification and node classification in protein-protein interaction graphs with minimal computational cost.
\end{abstract}

\section{Introduction}
\label{sec:introduction}

Deep learning has revolutionized computer vision by learning features directly from data. As a result deep neural networks have achieved state-of-the-art results on many difficult tasks such as image classification \cite{imagenet_cvpr09}, object detection \cite{lin2014microsoft}, object tracking \cite{muller2018trackingnet}, semantic segmentation \cite{cordts2016cityscapes}, depth estimation \cite{geiger2013vision} and activity understanding \cite{caba2015activitynet}, to name just a few examples. While there was a big emphasis on feature engineering before deep learning, the focus has now shifted to architecture engineering. In particular many novel architectures have been proposed, such as LeCun \cite{lecun1998gradient}, AlexNet \cite{krizhevsky2012imagenet}, VGG \cite{simonyan2014very}, GoogLeNet \cite{GoogLeNet2015}, ResNet \cite{he2016deep}, DenseNet \cite{huang2017densely}, ResNeXt \cite{xie2017aggregated} and SENet \cite{hu2018squeeze}. Results on each of the above mentioned tasks keep improving every year by innovations in architecture design. In essence, the community has shifted from feature engineering to architecture engineering.

\begin{figure}[!tp]
    \centering
    \begin{tabular}{cc}
    \includegraphics[trim=30mm 38mm 100mm 10mm, clip, width=0.9\columnwidth]{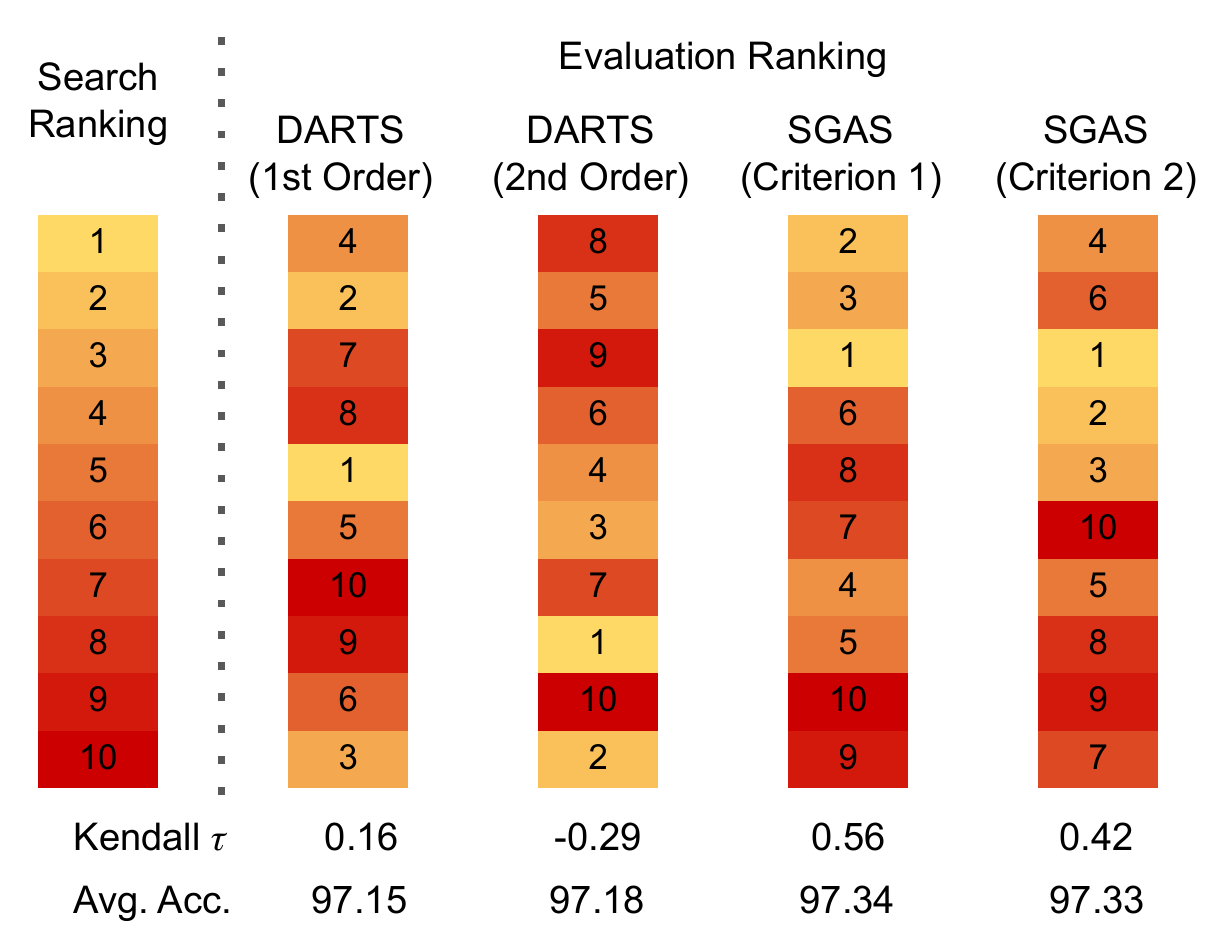}
    \end{tabular}
    \caption{\textbf{Comparison of search-evaluation Kendall $\tau$ coefficients.} We show Kendall $\tau$ correlations for architecture rankings between the search and the evaluation phase of DARTS and SGAS. Architectures are obtained from $10$ independent search runs.
    }
    \label{fig:intro_fig}
\end{figure}

In recent years, many efforts have been made to reduce the manual intervention required to obtain better models for a particular task. As a matter of fact, a new area of research, commonly referred to as meta-learning, has emerged in order to tackle such problems. The idea of meta-learning is to leverage prior experience in order to quickly find good algorithm configurations, network architectures and any required parameters for a new learning task. Examples of recent meta-learning approaches include automatic hyper-parameter search \cite{franceschi2018bilevel}, data-augmentation \cite{cubuk2018autoaugment}, finding novel optimizers \cite{andrychowicz2016learning} and architecture search \cite{zoph2016neural}. In particular, architecture search has sparked a lot of interest in the community. 
In this task, the search space is huge and manual search is prohibitive.

Early work by Zoph \etal \cite{zoph2016neural}, based on Reinforcement Learning, has shown very promising results. However, its high computational cost has prevented widespread adoption. Recently, differentiable architecture search (DARTS) \cite{liu2018darts} has been proposed as an alternative which makes architecture search differentiable and much more efficient. This has opened up a path towards computationally feasible architecture search. However, despite their success, current approaches still have a lot of limitations. During the search phase, network architectures are usually constructed from basic building blocks and evaluated on a validation set. Due to computational cost, the size of considered architectures is limited. In the evaluation phase, the best building blocks are used to construct larger architectures and they are evaluated on the test set.  As a result there is a large discrepancy between the validation accuracy during search and the test accuracy during evaluation. In this work, we propose a novel greedy architecture search algorithm, SGAS, which addresses this discrepancy and searches very efficiently.

\mysection{Contributions}  
Our contributions can be summarized as the following: \textbf{(1)} We propose SGAS, a greedy approach for neural architecture search with high correlation between the validation accuracy during the search phase and the final evaluation accuracy. \textbf{(2)} Our method discovers top-performing architectures with much less search cost than previous state-of-the-art methods such as DARTS. \textbf{(3)} Our proposed method is able to search architectures for both CNNs and GCNs across various datasets and tasks.

\section{Related Work}
\label{sec:related}
In the past, considerable success was achieved with hand-crafted architectures. One of the earliest successful architectures was LeNet \cite{lecun1998gradient}, a very simple convolutional neural network for optical character recognition. Other prominent networks include AlexNet \cite{krizhevsky2012imagenet}, VGG \cite{simonyan2014very} and GoogLeNet \cite{GoogLeNet2015} which revolutionized computer vision by outperforming all previous approaches in the ImageNet \cite{imagenet_cvpr09} challenge by a large margin. ResNet \cite{he2016deep} and DenseNet \cite{huang2017densely} were further milestones in architecture design. They showed the importance of residual and dense connections for designing very deep networks, an insight that influences modern architecture design to this day. 
Until recently, architecture innovations were a result of human insight and experimentation. The first successful attempts for architecture search were using reinforcement learning \cite{zoph2016neural} and evolutionary algorithms \cite{real2019regularized}. These works were extended with NASNet \cite{zoph2018learning} where a new cell-based search space and regularization technique were proposed. Another extension, ENAS \cite{pham2018efficient}, represents the entire search space as a single directed acyclic graph. A controller discovers architectures by searching for subgraphs that maximize the expected reward on the validation set. This setup allows for parameter sharing between child models making search very efficient. Further, PNAS \cite{liu2018progressive} introduced a sequential model-based optimization (SMBO) strategy in order to search for structures of increasing complexity. PNAS needs to evaluate $5$ times less models and reduces the computational cost by a factor of $8$ compared to NASNet. Yet, PNAS still requires thousands of GPU hours. 
One shot approaches \cite{brock2017smash,bender2018understanding,cai2018proxylessnas} further reduce the search time by training a single over-parameterized network with inherited/shared weights.
In order to search in a continuous domain \cite{saxena2016convolutional,ahmed2017connectivity,shin2018differentiable,veniat2018learning}, DARTS \cite{liu2018darts} proposes a continuous relaxation of the architecture representation,
making architecture search differentiable and hence much more efficient. As a result, DARTS is able to find good convolutional architectures at a fraction of the computational cost making NAS broadly accessible. Owed to the large success of DARTS, several extensions have been proposed recently. 
SNAS \cite{xie2018snas} optimizes parameters of a joint distribution for the search space in a cell. The authors propose a search gradient which optimizes the same objective as RL-based NAS, but leads to more efficient structural decisions.
P-DARTS \cite{chen2019progressive} attempts to overcome the depth gap issue between search and evaluation. This is accomplished by increasing the depth of searched architectures gradually during the training procedure.
PC-DARTS \cite{xu2019pc} leverages the redundancy in network space and only samples a subset of channels in super-net during search to reduce computation.

\section{Methodology} 
\label{sec:methodology}

\begin{figure*}[t]
    \vspace{-10pt}
    \centering
    \includegraphics[trim=0mm 5mm 0mm 5mm, clip, width=\textwidth]{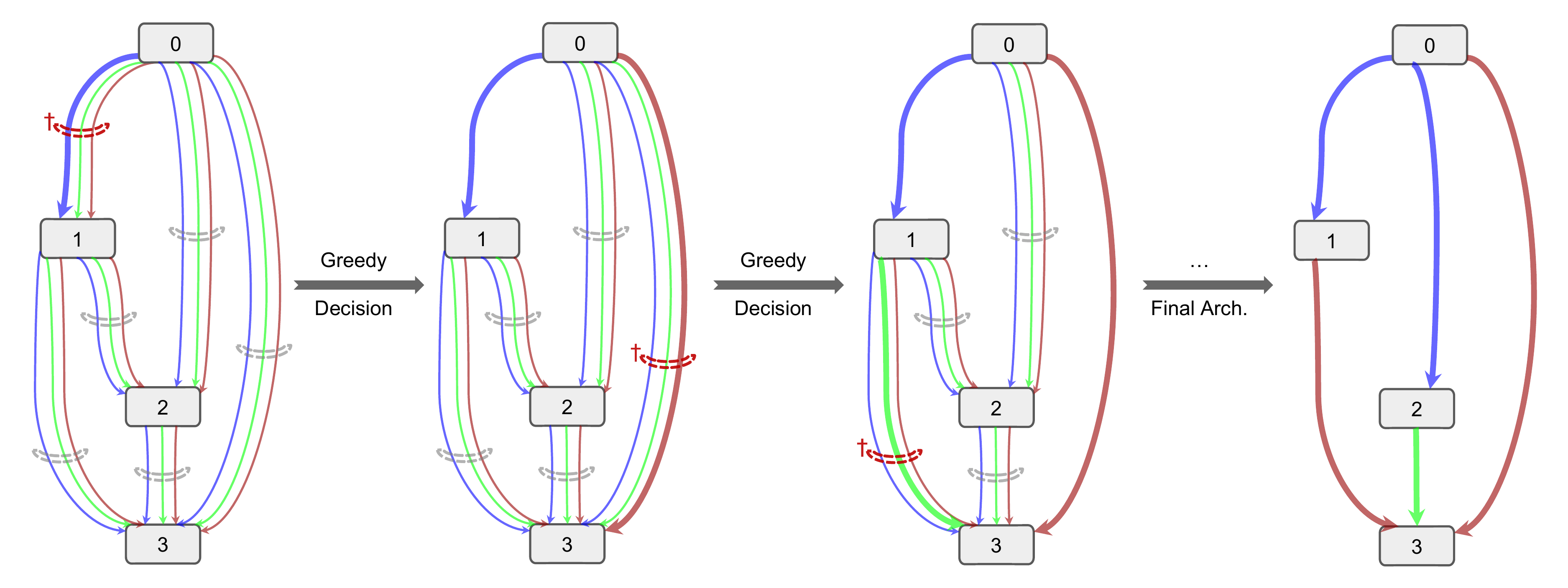}
    \caption{\textbf{Illustration of Sequential Greedy Architecture Search.} At each greedy decision epoch, an edge $(i^{\dagger}, j^{\dagger})$ is selected based on the selection criterion. A greedy decision will be made for the edge $(i^{\dagger}, j^{\dagger})$ by replacing $\bar{o}^{(i^\dagger,j^\dagger)}$ with $o^{(i^\dagger,j^\dagger)} = \mathrm{argmax}_{o \in \mathcal{O}} \enskip \alpha^{(i^\dagger,j^\dagger)}_o$. The corresponding architectural parameter $\alpha^{(i^{\dagger}, j^{\dagger})}$ will be removed from the bi-level optimization. Operations which were not chosen in a mixture operation will be pruned. At the end of the search phase, a stand-alone architecture without weight sharing will be obtained.}
\label{fig:pipeline}
\end{figure*}

\subsection{Preliminary - DARTS} \label{subsec:preliminary}
By reducing the search problem to searching for the best cell structure, cell-based NAS methods \cite{zoph2018learning,liu2018progressive,real2019regularized} are able to learn scalable and transferable architectures. The networks are composed of layers with identical cell structure but different weights. A cell is usually represented as a directed acyclic graph (DAG) with $N$ nodes including two input nodes, several intermediate nodes and a single output node. Each node is a latent representation denoted as $x^{(i)}$, where $i$ is its topological order in the DAG. Each directed edge $(i, j)$ in the DAG is associated with an operation $o^{(i, j)}$ that transfers the information from node $x^{(i)}$ to node $x^{(j)}$. In \textit{Differentiable Architecture Search} (DARTS) \cite{liu2018darts} and its variants \cite{xie2018snas,chen2019progressive,xu2019pc,he2020MiLeNAS}, the optimal architecture is derived from a discrete search space by relaxing the selection of operations to a continuous optimization problem. During the search phase, the operation of each edge is parameterized by architectural parameters $\alpha^{(i, j)}$ as a softmax mixture over all the possible operations within the operation space $\mathcal{O}$, \ie
$
	\bar{o}^{(i,j)}(x^{(i)}) = \sum_{o \in \mathcal{O}} \frac{\exp(\alpha_o^{(i,j)})}{\sum_{o' \in \mathcal{O}} \exp(\alpha_{o'}^{(i,j)})} o(x^{(i)})
$.
The input nodes are represented by the outputs from the previous two cells. Each intermediate node aggregates information flows from all of its predecessors,
$
	x^{(j)} = \sum_{i<j} \bar{o}^{(i, j)}(x^{(i)})
$.
The output node is defined as a concatenation of a fixed number of its predecessors. The learning procedure of architectural parameters involves a bi-level optimization problem:
\begin{align}
    	\min_\mathcal{A} \quad & \mathcal{L}_{val}(\mathcal{W}^*(\mathcal{A}), \mathcal{A}) \label{eq:outer} \\
    	\text{s.t.} \quad &\mathcal{W}^*(\mathcal{A}) = \mathrm{argmin}_\mathcal{W} \enskip \mathcal{L}_{train}(\mathcal{W}, \mathcal{A}) \label{eq:inner}
\end{align}
$\mathcal{L}_{train}$ and $\mathcal{L}_{val}$ denote the training and validation loss respectively. Owing to the continuous relaxation, the search is realized by optimizing a supernet. $\mathcal{W}$ is the set of weights of the supernet and $\mathcal{A}$ is the set of the architectural parameters. DARTS \cite{liu2018darts} proposed to solve this bi-level problem by a first/second order approximation. At the end of the search, the final architecture is derived by selecting the operation with highest weight for every mixture operation,
$
	o^{(i,j)} = 
	\mathrm{argmax}_{o \in \mathcal{O}} \enskip \alpha^{(i,j)}_o
$.

\begin{algorithm*}[t]
\DontPrintSemicolon
Create architectural parameters $\mathcal{A} = \{\alpha^{(i, j)}\}$ and supernet weights $\mathcal{W}$\;
Create a mixed operation $\bar{o}^{(i,j)}$ parameterized by $\alpha^{(i,j)}$ for each edge $(i,j)$\;
\While{not terminated}{
	1. Update undetermined architecture parameters $\mathcal{A}$ by descending
	$\nabla_{\mathcal{A}} \mathcal{L}_{val}(\mathcal{W}, \mathcal{A})$\;
	2. Update weights $\mathcal{W}$ by descending $\nabla_{\mathcal{W}} \mathcal{L}_{train}(\mathcal{W}, \mathcal{A})$\;
	\quad (since the weights of unchosen operations are pruned, only the remaining weights need to be updated)\;
	3. If a decision epoch, select an edge $(i^\dagger,j^\dagger)$ based on the greedy \textit{Selection Criterion}
	\\\quad Determine the operation by replacing $\bar{o}^{(i^\dagger,j^\dagger)}$ with $o^{(i^\dagger,j^\dagger)} = \mathrm{argmax}_{o \in \mathcal{O}} \enskip \alpha^{(i^\dagger,j^\dagger)}_o$
	\\\quad Prune unchosen weights from $\mathcal{W}$, Remove $\alpha^{(i^\dagger, j^\dagger)}$ from $\mathcal{A}$\;
}
Derive the final architecture based on chosen operations
\caption{{\sc SGAS} -- Sequential Greedy Architecture Search}
\label{algo:pseudocode}
\end{algorithm*}

\subsection{Search-Evaluation Correlation}
A popular pipeline of existing NAS algorithms \cite{zoph2018learning,liu2018darts} includes two stages: a search phase and an evaluation phase. In order to reduce computational overhead, previous works \cite{zoph2018learning,liu2018darts} first search over a pre-defined search space with a lightweight proxy model on a small proxy dataset. After the best architecture cell/encoding is obtained, the final architecture is built and trained from scratch on the target dataset. This requires that the true performance during evaluation can be inferred during the search phase. However, this assumption usually does not hold due to the discrepancy in dataset, hyper-parameters and network architectures between the search and evaluation phases. The best ranking derived from the search phase does not imply the actual ranking in the final evaluation. In practice, the correlation between the performances of derived architectures during the search and evaluation phases is usually low. In this paper, we refer to this issue as \textbf{\secor}. Recent work by Sciuto \etal \cite{sciuto2019evaluating} also analyzes this issue and suggests that the Kendall $\tau$ metric \cite{kendall1938new} could be used to evaluate the search phase. They show that the widely used \emph{weight sharing} technique actually decreases the correlation. The Kendall $\tau$ metric \cite{kendall1938new} is a common measurement of the correlation between two rankings. The Kendall $\tau$ coefficient can be computed as 
$
	\tau = \frac{N_{c} - N_{d}}{\frac{1}{2}n(n-1)}
$, where $N_{c}$ and $N_{d}$ are the number of concordant pairs and the number of discordant pairs respectively. It is a number in the range from $-1$ to $1$ where $-1$ corresponds to a perfect negative correlation and $1$ to a perfect positive correlation. If the Kendall $\tau$ coefficient is $0$, the rankings are completely independent. An ideal NAS method should have a high \textbf{search-evaluation Kendall $\tau$ coefficient}. We take DARTS \cite{liu2018darts} as an example and show its Kendall $\tau$ in \figLabel \ref{fig:intro_fig}. It is calculated between the rankings during search phase and evaluation phase. The rankings are determined according to the validation accuracy and the final evaluation accuracy after 10 different runs on the CIFAR-10 dataset. The Kendall $\tau$ coefficients for DARTS are only $0.16$ and $-0.29$ for the $1$st-order and $2$nd-order versions respectively. Therefore, it is impossible to make reliable predictions regarding the final test accuracy based on the search phase.

\subsection{Sequential Greedy Architecture Search}
In order to alleviate the \textbf{\secor} problem, the core aspects are to reduce $\textbf{(1)}$ the discrepancy between the search and evaluation phases and $\textbf{(2)}$ the negative effect of weight sharing. We propose to solve the bi-level optimization (\eqLabel \ref{eq:outer}, \ref{eq:inner}) in a sequential greedy fashion to reduce the model discrepancy and the weight sharing progressively. As mentioned in \secLabel \ref{subsec:preliminary}, DARTS-based methods \cite{liu2018darts,chen2019progressive,xu2019pc} solve the relaxed problem fully and obtain all the selected operations at the end. Instead of solving the complete problem directly, we divide it into sub-problems and solve them sequentially with a greedy algorithm. The sub-problems are defined based on the directed edges in the DAG. We pick the operation for edges greedily in a sequential manner and solve the remaining sub-problem iteratively. The iterative procedure is shown in \algLabel \ref{algo:pseudocode}. At each decision epoch, we choose one edge $(i^{\dagger}, j^{\dagger})$ according to a pre-defined \textit{selection criterion}. A greedy optimal choice is made for the selected edge by replacing the corresponding mixture operation $\bar{o}^{(i^\dagger,j^\dagger)}$ with $o^{(i^\dagger,j^\dagger)} = \mathrm{argmax}_{o \in \mathcal{O}} \enskip \alpha^{(i^\dagger,j^\dagger)}_o$. The architectural parameters $\alpha^{(i^\dagger,j^\dagger)}$ and the weights of the remaining paths within the mixture operations are no longer needed; we prune and exclude them from the latter optimization. As a side benefit, the efficiency improves as parameters in $\mathcal{A}$ and $\mathcal{W}$ are pruned gradually in the optimization loop. The search procedure of the remaining $\mathcal{A}$ and $\mathcal{W}$ forms a new sub-problem which will be solved iteratively. At the end of the search phase, a \textit{stand-alone network without weight sharing is obtained}, as illustrated in \figLabel \ref{fig:pipeline}. Therefore, the model discrepancy is minimized and the validation accuracy during the search phase reflects the final evaluation accuracy much better. To maintain the optimality, the design of the \textit{selection criterion} is crucial. We consider three aspects of edges which are the \textit{edge importance}, the \textit{selection certainty} and the \textit{selection stability}.

\mysection{Edge Importance} Similar to DARTS \cite{liu2018darts}, a \textit{zero} operation is included in the search space to indicate a lack of connection. Edges that are important should have a low weight in the \textit{zero} operation. Thus, the edge importance is defined as the summation of weights over non-zero operations:
\begin{align}
S_{EI}^{(i,j)} = \sum_{o \in \mathcal{O}, o \neq zero} \frac{\exp(\alpha_{o}^{(i, j)})}{\sum_{o' \in \mathcal{O}}\exp(\alpha_{o'}^{(i, j)})}
\label{eq:ei}
\end{align}
\mysection{Selection Certainty} Entropy is a common measurement of uncertainty of a distribution. The normalized softmax weights of non-zero operations can be regarded as a distribution, $p_{o}^{(i,j)} = \frac{\exp(\alpha_{o}^{(i, j)})}{S_{EI}^{(i,j)}\sum_{o' \in \mathcal{O}}\exp(\alpha_{o'}^{(i, j)})}, o \in \mathcal{O}, o \neq zero$. We define the \textit{selection certainty} as the \textit{complement} of the normalized entropy of the operation distribution:
\begin{align}
S_{SC}^{(i,j)} = 1 -  \frac{-\sum_{o \in \mathcal{O}, o \neq zero} p_{o}^{(i,j)} \log(p_{o}^{(i,j)})}{\log(|\mathcal{O}|-1)} 
\label{eq:sc}
\end{align}
\mysection{Selection Stability} In order to incorporate the history information, we measure the movement of the operation distribution. Kullback–Leibler divergence and histogram intersection \cite{swain1991color} are two popular methods to detect changes in distribution. For simplicity, we choose the latter one. The average selection stability at step $T$ with a history window size $K$ is computed as follows:
\begin{align}
S_{SS}^{(i,j)} = \frac{1}{K} \sum_{t=T-K}^{T-1} \sum_{o_{t} \in \mathcal{O} o_{t}, \neq zero} \min (p_{o_{t}}^{(i,j)}, p_{o_{T}}^{(i,j)})
\label{eq:ss}
\end{align}
In our experiments, we consider the following two criteria:

\mysection{Criterion 1} An edge $(i^\dagger,j^\dagger)$ with a high \textit{edge importance} $S_{EI}^{(i,j)}$ and a high \textit{selection certainty} $S_{SC}^{(i,j)}$ will be selected. We normalize $S_{EI}^{(i,j)}$ and $S_{SC}^{(i,j)}$, compute the final score and pick the edge with the highest score:
\begin{align}
S_{1}^{(i,j)} = \text{normalize}(S_{EI}^{(i,j)})*\text{normalize}(S_{SC}^{(i,j)})
\label{eq:c1}
\end{align}

\mysection{Criterion 2} In addition to \textit{Criterion 1}, we also consider that the selected edge $(i^\dagger,j^\dagger)$ should have a high \textit{selection stability}. The final score is defined as follows:
\begin{align}
S_{2}^{(i,j)} = S_{1}^{(i,j)}*\text{normalize}(S_{SS}^{(i,j)})
\label{eq:c2}
\end{align}
where $normalize(\cdot)$ denotes a standard Min-Max scaling normalization. For a fair comparison with existing works \cite{zoph2016neural,real2019regularized,liu2018darts}, two incoming edges are preserved for every intermediate node in the DAG. Once a node has two determined incoming edges, its other incoming edges will be pruned. We refer to our method as \textbf{Sequential Greedy Architecture Search} (SGAS). \figLabel \ref{fig:intro_fig} shows that SGAS with Criterion $1$ and $2$ improves the Kendall $\tau$ correlation coefficients to $0.56$ and $0.42$ respectively. As expected from the much higher search-evaluation correlation SGAS outperform DARTS in terms of average accuracy significantly.

\section{Experiments}
\label{sec:experiments}
We use our SGAS to automatically find architectures for both CNNs and GCNs.
The CNN architectures discovered by SGAS outperform the state-of-the-art (SOTA) in image classification on CIFAR-10 \cite{Krizhevsky2009LearningML} and ImageNet \cite{imagenet_cvpr09}. 
Similarly, the discovered GCN architectures outperform the state-of-the-art methods for point cloud classification on ModelNet \cite{20143Dmodelnet}
and node classification in biological graphs using the PPI \cite{Zitnik2017PredictingMFppi} dataset.

\subsection{Searching CNN architectures with SGAS}
\savespace
\subsubsection{Architecture Search on CIFAR-10}
 As is common practice, we first search for normal cells and reduction cells with a small network for image classification on CIFAR-$10$. CIFAR-$10$ is a small popular dataset containing $50K$ training images and $10K$ testing images. Then, a larger network is constructed by making necessary changes in channel size and stacking the searched cells multiple times. The larger network is retrained on CIFAR-$10$ to compare its performance with other state-of-the-art methods. Finally, we show the transferability of our SGAS by stacking even more cells and evaluating on ImageNet. We show that SGAS consistently achieves the top performance.

\mysection{Search Space} We keep our search space the same as DARTS, which has $8$ candidate operations: \textit{skip-connect}, \textit{max-pool-3$\times$3}, \textit{avg-pool-3$\times$3}, \textit{sep-conv-3$\times$3}, \textit{sep-conv-5$\times$5}, \textit{dil-conv-3$\times$3}, \textit{dil-conv-5$\times$5}, \textit{zero}. 
During the search phase, we stack $6$ normal cells and $2$ reduction cells to form a network. 
Two reduction cells are inserted at a network depth of $1/3$ and $2/3$ respectively.
The stride of each convolution in normal cells is $1$, so the spatial size of an input feature map does not change. In reduction cells, convolutions with stride $2$ are used to reduce the spatial resolution of feature maps.
There are $7$ nodes with $4$ intermediate nodes and $14$ edges in each cell during search. The first and second input nodes of the cell are set equal to the outputs of the two previous cells respectively. 
The output node of a cell is the depth-wise concatenation of all the intermediate nodes.

\mysection{Training Settings}
We keep the training setting the same as in DARTS \cite{liu2018darts}. A small network consisting of $6$ normal cells and $2$ reduction cells with an initial channel size $16$ is trained on CIFAR-$10$.
We perform architecture search for $50$ epochs with a batch size of $64$.
SGD is used to optimize the model weights $\mathcal{W}$ with an initial learning rate $0.025$, momentum $0.9$ and weight decay $3\times10^{-4}$. For architecture parameters $\mathcal{A}$, the Adam optimizer with an initial learning rate $3 \times 10^{-4}$, momentum $(0.5, 0.999)$ and weight decay $10^{-3}$ is used.
Instead of training the entire super-net throughout the search phase, SGAS makes decisions sequentially in a greedy fashion. 
After warming up for $9$ epochs, SGAS begins to select one operation for one selected edge every $5$ epochs using \textit{Criterion 1} or \textit{Criterion 2} as the selection criterion. For \textit{Criterion 2}, we set the history window size $K$ to 4.
The batch size is increased by $8$ after each greedy decision, which further boosts the searching efficiency of SGAS. We provide a thorough discussion and ablation study on the choices of hyper-parameters in the \supp. The search takes only $0.25$ day ($6$ hours) on a single NVIDIA GTX 1080Ti.

\begin{table*}
\vspace{-5pt}
\centering
\resizebox{0.58\linewidth}{!}{%
\begin{tabular}{lccccc}
\toprule
\textbf{{Architecture}} & \textbf{Test Err.} & \textbf{Params} & \textbf{Search Cost} & \textbf{{Search}} \\
&                            \textbf{ (\%)} & \textbf{(M)} & \textbf{(GPU-days)} & \textbf{{Method}}\\
\midrule
DenseNet-BC \cite{huang2017densely} & 3.46 & 25.6 & - & manual \\ 
\midrule
NASNet-A \cite{zoph2018learning} & 2.65 & 3.3 & 1800 & RL \\ 
AmoebaNet-A \cite{real2019regularized} & 3.34$\pm$0.06 & 3.2 & 3150 & evolution \\ 
AmoebaNet-B \cite{real2019regularized} & 2.55$\pm$0.05 & 2.8 & 3150 & evolution \\ 
Hier-Evolution \cite{liu2017hierarchical} & 3.75$\pm$0.12 & 15.7 & 300 & evolution \\ 
PNAS \cite{liu2018progressive} & 3.41$\pm$0.09 & 3.2 & 225 & SMBO \\ 
ENAS \cite{pham2018efficient} & 2.89 & 4.6 & 0.5 & RL \\ 
NAONet-WS \cite{luo2018neural} & 3.53 & 3.1 & 0.4 & NAO \\ 
\midrule
DARTS ($1^{\text{st}}$ order) \cite{liu2018darts} & 3.00$\pm$0.14 & 3.3 & 0.4 & gradient \\ 
DARTS ($2^{\text{nd}}$ order) \cite{liu2018darts} & 2.76$\pm$0.09 & 3.3 & 1 & gradient \\ 
SNAS (mild) \cite{xie2018snas} & 2.98 & 2.9 & 1.5 & gradient\\ 
ProxylessNAS \cite{cai2018proxylessnas} & 2.08 & - & 4 & gradient \\ 
P-DARTS \cite{chen2019progressive} & 2.5 & 3.4 & 0.3 & gradient \\ 
BayesNAS \cite{zhou2019bayesnas} & 2.81$\pm$0.04 & 3.4 & 0.2 & gradient \\ 
PC-DARTS \cite{xu2019pc} & 2.57$\pm$0.07 & 3.6 & 0.1 & gradient \\ 
\midrule
SGAS (Cri.1 avg.) & 2.66$\pm$$0.24^{*}$ & 3.7 & 0.25 & gradient \\ 
SGAS (Cri.1 best) & 2.39 & 3.8 & 0.25 & gradient \\ 
SGAS (Cri.2 avg.) & 2.67$\pm$$0.21^{*}$ & 3.9 & 0.25 & gradient \\ 
SGAS (Cri.2 best) & 2.44 & 4.1 & 0.25 & gradient \\ 
\bottomrule
\end{tabular}
}
\vspace{3pt}
\caption{\textbf{Performance comparison with state-of-the-art image classifiers on CIFAR-10.} We report the average and best performance of SGAS (Cri.1) and SGAS (Cri.2). \textit{Criterion 1} and \textit{Criterion 2} are used in the search respectively. *Note that mean and standard derivation are computed across 10 independently searched architectures.}
\savespace
\label{tbl:compare_cifar10}
\end{table*}

\savespace
\subsubsection{Architecture Evaluation on CIFAR-10}
We run $10$ \textit{independent searches} to get $10$ architectures with \textit{Criterion 1} or \textit{Criterion 2}, as shown in \figLabel \ref{fig:intro_fig}. To highlight the stability of the search method, we evaluate the discovered architectures on CIFAR-$10$ and report the mean and standard deviation of the test accuracy across those $10$ models and the performance of the best model in \tblLabel \ref{tbl:compare_cifar10}. It is important to mention that other related works in \tblLabel \ref{tbl:compare_cifar10} only report the mean and standard deviation for their \textit{best architecture} with different runs on evaluation. 

\mysection{Training Settings}
We train a large network of $20$ cells with a initial channel size $36$. The SGD optimizer is used during $600$ epochs with a batch size of $96$. The other hyper-parameters remain the same as the search phase. Cutout with length $16$, auxiliary towers with weight $0.4$ and path dropout with probability $0.3$ are used as in DARTS \cite{liu2018darts}.

\mysection{Evaluation Results and Analysis}
We compare our results with other methods in \tblLabel \ref{tbl:compare_cifar10} and report the average and best performance for both \textit{Criterion 1} and \textit{Criterion 2}. We outperform our baseline DARTS by a significant margin with test errors of $2.39\%$ and $2.44\%$ respectively while only using $0.25$ day ($6$ hours) on a single NVIDIA GTX 1080Ti.

\savespace
\subsubsection{Architecture Evaluation on ImageNet}
The architecture evaluation on ImageNet uses the cell architectures that we obtained after searching on CIFAR-10.

\mysection{Training Settings}
We choose the $3$ best performing cell architectures on CIFAR-10 for each \textit{Criterion} and train them on ImageNet. For this evaluation, we build a large network with $14$ cells and $48$ initial channels and train for $250$ epochs with a batch size of $1024$. The SGD optimizer with an initial learning rate of $0.5$, a momentum of $0.9$ and a weight decay of $3\times10^{-5}$ is used. We run these experiments on $8$ Nvidia Tesla V100 GPUs for three days.

\begin{figure*}[h]
 \vspace{-30pt}
 \centering
 \begin{tabular}{@{}c@{\hspace{1mm}}c@{\hspace{1mm}}c@{}}
 
 		\includegraphics[trim=10mm 13mm 0mm 13mm, clip, width=\columnwidth]{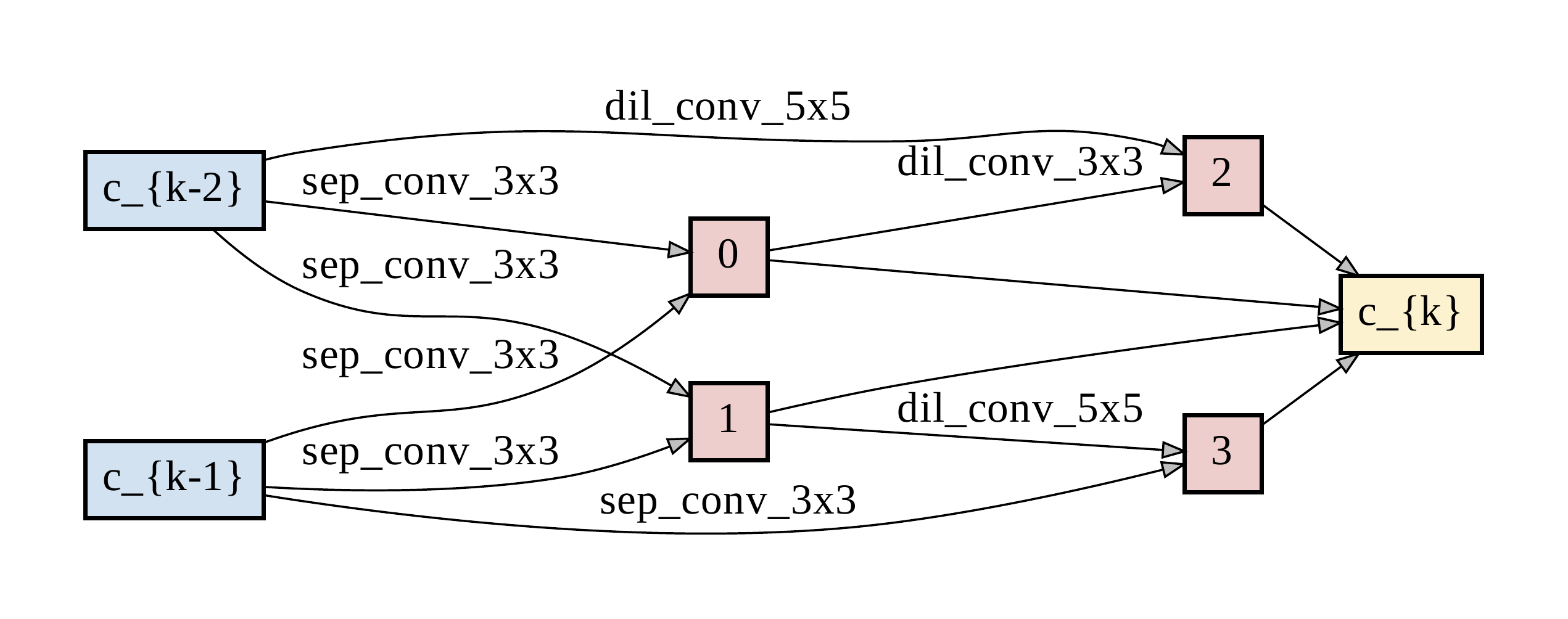} &
 		\includegraphics[trim=0mm 13mm 10mm 13mm, clip, height=0.35\columnwidth]{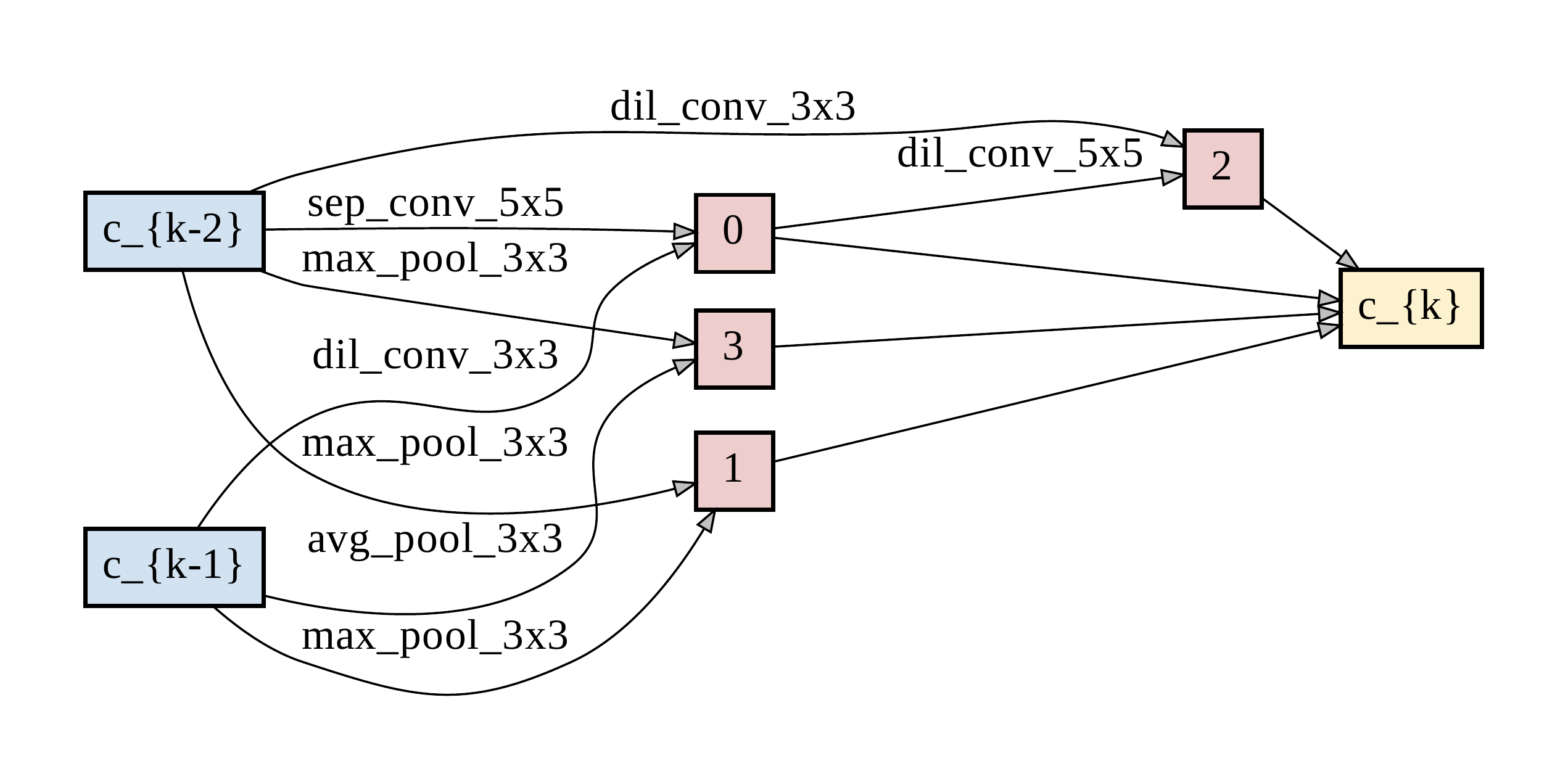} \\
		
 		\small (a) Normal cell &
 		\small (b) Reduction cell \\
		
 \end{tabular}
 \caption{\textbf{Best cell architecture on Imagenet with SGAS \textit{Crit. 2}}}
 \label{fig:visualizations_cnn}
 \vspace{-10pt}
 \end{figure*}

\mysection{Evaluation Results and Analysis}
In \tblLabel \ref{tbl:compare_imagenet} we compare our models with SOTA hand-crafted architectures (manual) and models obtained through other search methods. We apply the mobile setting for ImageNet, which has an image size of $224\times224$ and restricts the number of multi-add operations to $600M$. Our best performing models SGAS (Cri.1 best) and SGAS (Cri.2 best) outperform all the other methods with top-1 errors $24.2\%$ and $24.1\%$ respectively while using only a search cost of $0.25$ GPU day on one NVIDIA GTX 1080Ti. SGAS (Cri.2) outperforms SGAS (Cri.1) showing the effectiveness of integrating \textit{selection stability} into the selection criterion. The best performing cells of SGAS (Cri.2 best) are illustrated in \figLabel \ref{fig:visualizations_cnn}.
\begin{table*}
\vspace{-10pt}
\centering
\resizebox{0.70\linewidth}{!}{%
\begin{tabular}{lcccccc}
\toprule
\textbf{{Architecture}} & \multicolumn{2}{c}{\textbf{Test Err. (\%)}} & \textbf{Params} & $\times+$ & \textbf{Search Cost} & \textbf{{Search}} \\
&                            \textbf{top-1} & \textbf{top-5} & \textbf{(M)} & \textbf{(M)} & \textbf{(GPU-days)} & \textbf{Method}\\
\midrule
Inception-v1 \cite{GoogLeNet2015} & 30.2 & 10.1 & 6.6 & 1448 & - & manual \\ 
MobileNet \cite{howard2017mobilenets}& 29.4 & 10.5 & 4.2 & 569 & - & manual \\ 
ShuffleNet 2x (v1) \cite{zhang2018shufflenet} & 26.4 & 10.2 & $\sim$5 & 524 & - & manual \\ 
ShuffleNet 2x (v2) \cite{ma2018shufflenet} & 25.1 & - & $\sim$5 & 591 & - & manual \\ 
\midrule
NASNet-A \cite{zoph2018learning} & 26 & 8.4 & 5.3 & 564 & 1800 & RL \\ 
NASNet-B \cite{zoph2018learning} & 27.2 & 8.7 & 5.3 & 488 & 1800 & RL \\ 
NASNet-C \cite{zoph2018learning} & 27.5 & 9 & 4.9 & 558 & 1800 & RL \\ 
AmoebaNet-A \cite{real2019regularized} & 25.5 & 8 & 5.1 & 555 & 3150 & evolution \\ 
AmoebaNet-B \cite{real2019regularized} & 26 & 8.5 & 5.3 & 555 & 3150 & evolution \\ 
AmoebaNet-C \cite{real2019regularized} & 24.3 & 7.6 & 6.4 & 570 & 3150 & evolution \\ 
FairNAS-A \cite{chu2019fairnas} & 24.7 & 7.6 & 4.6 & 388 & 12 & evolution \\
PNAS \cite{liu2018progressive} & 25.8 & 8.1 & 5.1 & 588 & 225 & SMBO \\ 
MnasNet-92 \cite{tan2019mnasnet} & 25.2 & 8 & 4.4 & 388 & - & RL \\
\midrule
DARTS ($2^{\text{nd}}$ order) \cite{liu2018darts} & 26.7 & 8.7 & 4.7 & 574 & 4.0 & gradient \\ 
SNAS (mild) \cite{xie2018snas} & 27.3 & 9.2 & 4.3 & 522 & 1.5 & gradient \\ 
ProxylessNAS \cite{cai2018proxylessnas} & 24.9 & 7.5 & 7.1 & 465 & 8.3 & gradient \\ 
P-DARTS \cite{chen2019progressive} & 24.4 & 7.4 & 4.9 & 557 & 0.3 & gradient \\ 
BayesNAS \cite{zhou2019bayesnas} & 26.5 & 8.9 & 3.9 & - & 0.2 & gradient \\ 
PC-DARTS \cite{xu2019pc} & 25.1 & 7.8 & 5.3 & 586 & 0.1 & gradient \\
\midrule
SGAS (Cri.1 avg.) & 24.41$\pm$0.16 & 7.29$\pm$0.09 & 5.3 & 579 & 0.25 & gradient \\
SGAS (Cri.1 best) & 24.2 & 7.2 & 5.3 & 585 & 0.25 & gradient \\ 
SGAS (Cri.2 avg.) & 24.38$\pm$0.22 & 7.39$\pm$0.07 & 5.4 & 597 & 0.25 & gradient \\ 
SGAS (Cri.2 best) & \textbf{24.1} & 7.3 & 5.4 & 598 & 0.25 & gradient \\ 
\bottomrule
\end{tabular}
}
\vspace{3pt}
\caption{\textbf{Comparison with state-of-the-art classifiers on ImageNet.} We transfer the top $3$ performing architectures on CIFAR-10 to ImageNet in the mobile setting. $\times+$ denote multiply-add operations.
The average and best performance of SGAS are reported.}
\savespace
\label{tbl:compare_imagenet}
\end{table*}

\subsection{Searching GCN architectures with SGAS}
Recently, GCNs have achieved impressive performance on point cloud segmentation \cite{Li2019DeepGCNs}, biological graph node classification \cite{Li2019DeepGCNsMG} and video recognition \cite{xu2019gtad} by training DeepGCNs \cite{Li2019DeepGCNs,Li2019DeepGCNsMG}.
However, this hand-crafted architecture design requires adequate effort by an human expert. 
The main component of DeepGCNs is the GCN backbone. We explore an automatic way to design the GCN backbone using SGAS.
Our backbone network is formed by stacking the graph convolutional cell discovered by SGAS. Our GCN cell consists of 6 nodes. We use fixed $1 \times 1$ convolutions in the first two nodes, and set the input to them equal to the output from the previous two layers.
Our experiments on GCNs have two stages. First, we apply SGAS to search for the graph convolutional cell using a small dataset and obtain $10$ architectures from $10$ runs. Then, $10$ larger networks are constructed by stacking each discovered cell multiple times. 
The larger networks are trained on the same dataset or a larger one to evaluate their performance. We report the best and average performance of these $10$ architectures.
We show the effectiveness of SGAS in GCN architecture search by comparisons with SOTA hand-crafted methods and Random Search. 

\savespace
\subsubsection{Architecture Search on ModelNet10}
ModelNet \cite{20143Dmodelnet} is a dataset for 3D object classification with two variants, ModelNet10 and ModelNet40 containing objects from $10$ and $40$ classes respectively. We conduct GCN architecture search on ModelNet10 and then evaluate the final performance on ModelNet40.

\mysection{Search Space}
Our graph convolutional cell has $10$ candidate operations:
\textit{conv-1$\times$1}, \textit{MRConv} \cite{Li2019DeepGCNs}, \textit{EdgeConv} \cite{dgcnn}, \textit{GAT} \cite{velivckovic2017graph}, \textit{SemiGCN} \cite{Kipf2016SemiSupervisedCW}, \textit{GIN} \cite{Xu2018HowPAGIN}, \textit{SAGE} \cite{hamilton2017inductive}, \textit{RelSAGE}, \textit{skip-connect}, and \textit{zero} operation.
Please refer to our \supp for more details of these GCN operators.
We use \textit{k nearest neighbor} in the first operation of each cell to construct edges (we use $k = 9$ by default unless it is specified). These edges are then shared in the following operations inside the cell.
Dilated graph convolutions with the same linearly increasing dilation rate schedule as proposed in DeepGCNs \cite{Li2019DeepGCNs} are applied to the cells.

\mysection{Training Settings}
We sample $1024$ points from the 3D models in ModelNet10.
We use $2$ cells with $32$ initial channels and search the architectures for $50$ epochs with batch size $28$. 
SGD is used to optimize the model weights with initial learning rate $0.005$, momentum $0.9$ and weight decay $3\times10^{-4}$. The Adam optimizer with the same parameters as in the search for CNNs is used to optimize architecture parameters.
After warming up for $9$ epochs, SGAS begins to select one operation for a selected edge every $7$ epochs.
We experimented with both selection criteria, \textit{Criterion 1} and \textit{Criterion 2}. We use a history window of $4$ for \textit{Cri.2}. The batch size increases by $4$ after each decision. The search takes around $0.19$ GPU day on one NVIDIA GTX 1080Ti.

\begin{table}
\small
\setlength{\tabcolsep}{6pt}
\centering
\resizebox{\columnwidth}{!}{%
\begin{tabular}{lccc}
\toprule
\textbf{Architecture} & \textbf{OA} & \textbf{Params} & \textbf{Search Cost} \\ 
&                           \textbf{(\%)} & \textbf{(M)} & \textbf{(GPU-days)}\\
\midrule
3DmFV-Net \cite{ben20183dmfv} & 91.6 & 45.77 & manual \\ 
SpecGCN \cite{Wang_2018_ECCV} & 91.5 & 2.05 & manual \\ 
PointNet++ \cite{qi2017pointnet++} & 90.7 & 1.48 & manual \\ 
PCNN \cite{Atzmon:2018:PCN:3197517.3201301} & 92.3 & 8.2 & manual \\ 
PointCNN \cite{Li2018PointCNNCO} & 92.2 & 0.6 & manual \\ 
DGCNN \cite{dgcnn} & 92.2 & 1.84 & manual \\ 
KPConv \cite{thomas2019KPConv} & 92.9 & 14.3 & manual \\ 
\midrule
Random Search & 92.65$\pm$0.33 & 8.77 & random \\ 
SGAS (Cri.1 avg.) & 92.69$\pm$0.20 & 8.78 & 0.19 \\ 
SGAS (Cri.1 best) & 92.87 & 8.63 & 0.19 \\ 
SGAS (Cri.2 avg.) & 92.93$\pm$0.19 & 8.87 & 0.19 \\ 
SGAS (Cri.2 best) & \textbf{93.23} & 8.49 & 0.19 \\
SGAS (Cri.2 small best) & 93.07 & 3.86 & 0.19 \\

\bottomrule
\end{tabular}
}
\vspace{3pt}
\caption{\textbf{Comparison with state-of-the-art architectures for 3D object classification on ModelNet40.} $10$ architectures are derived for both SGAS and Random Search within the same search space.
}
\savespace
\label{tbl:modelnet40}
\end{table}

\savespace
\subsubsection{Architecture Evaluation on ModelNet40}
After searching for $10$ architectures on ModelNet10, we form a large backbone network for each and train them on ModelNet40. The performance of 3D point cloud classification is evaluated with the overall accuracy (OA). We also apply Random Search to the same search space to obtain $10$ architectures as our random search baseline.

\mysection{Training Settings}
We stack the searched cell $9$ times with channel size $128$. 
We also form small networks by stacking the cell $3$ times with the same channel size.
We use $k=20$ for all the large networks and $k=9$ for the small ones.
Adam is used to optimize the weights with initial learning rate $0.001$ and weight decay $1\times10^{-4}$.
We sample $1024$ points as input. 
Our architectures are all trained for $400$ epochs with batch size of $32$. We report the mean and standard deviation of the accuracy on the test dataset of the $10$ discovered architectures; we also report the accuracy of the best performing model of the big and the small networks. 

\begin{figure*}
 \centering
 \begin{tabular}{@{}c@{\hspace{1mm}}c@{\hspace{1mm}}c@{}}
 		\includegraphics[trim=0mm 5mm 0mm 10mm, clip, width=\columnwidth]{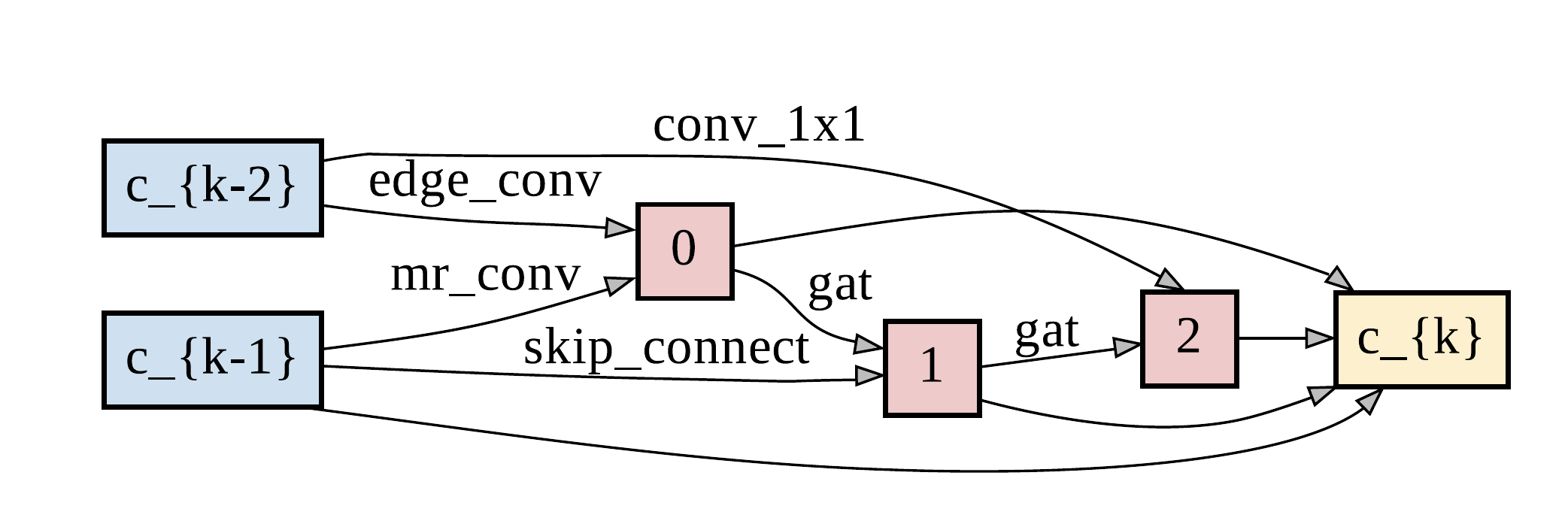} &
 		\includegraphics[trim=0mm 10mm 0mm 10mm, clip, width=\columnwidth]{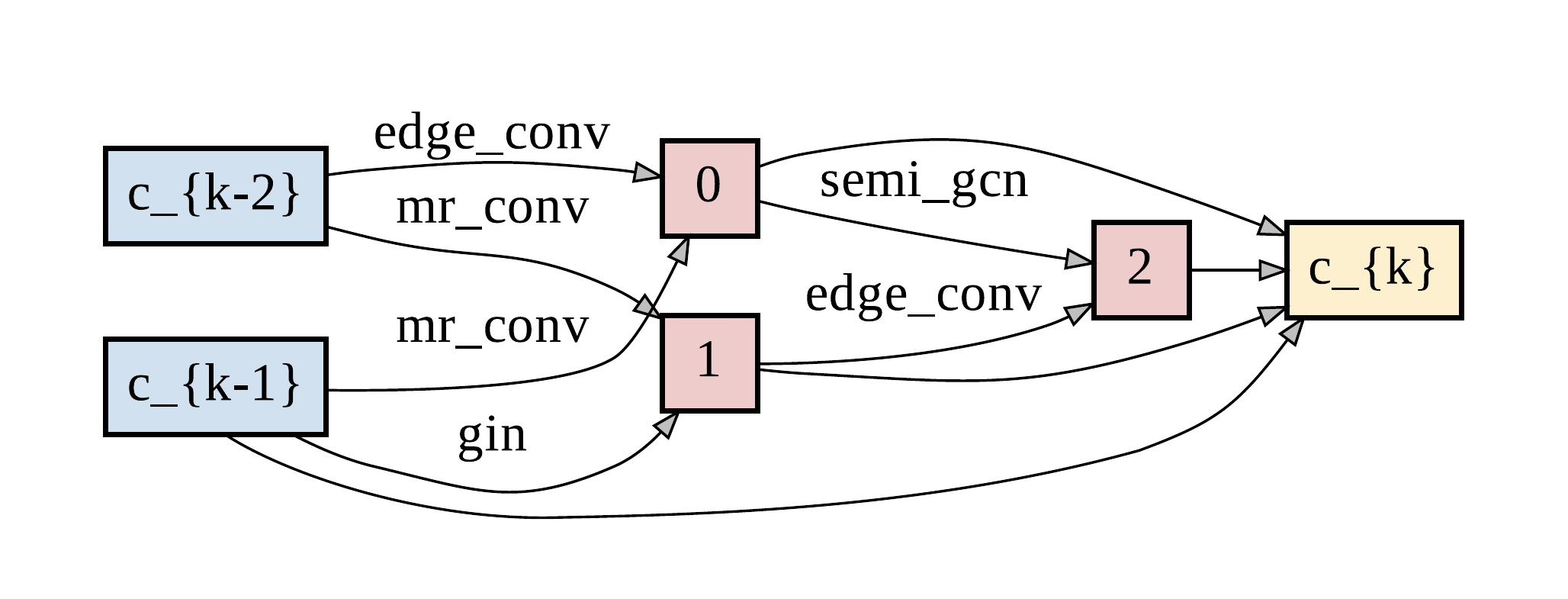}\\
		
		\small (a) Normal cell of the best model with SGAS \textit{Crit. 2} on ModelNet40 &
		\small (b) Normal cell of the best model with SGAS \textit{Crit. 1} on PPI\\

 \end{tabular}
 \caption{\textbf{Best cell architectures on ModelNet40 and PPI}}
 \savespace
 \label{fig:visualizations_gcn}
 \end{figure*}

\mysection{Evaluation Results and Analysis}
We compare the performance of our discovered architectures with SOTA hand-crafted methods and architectures obtained by Random Search for 3D point clouds classification on ModelNet40.
\tblLabel \ref{tbl:modelnet40} shows that SGAS (Cri.2 best), the best architecture discovered by our SGAS with \textit{Criterion 2}, outperforms all the other models. 
The smaller network SGAS (Cri.2 small best) discovered by SGAS with \textit{Criterion 2} also outperforms all the hand-crafted architectures. 
Owing to a well-designed search space, Random Search is a strong baseline. The performance of SGAS surpasses the hand-crafted architectures and Random Search, demonstrating the effectiveness of SGAS for GCN architecture search. 
The best architecture for this task can be found in \figLabel \ref{fig:visualizations_gcn} (a).

\savespace
\subsubsection{Architecture Search on PPI}
PPI is a popular biological graph dataset in the data mining domain. We search for GCN architectures on the PPI dataset for the task of node classification.

\mysection{Training Settings}
We use $1$ cell with $32$ channels. We train and search the architectures for $50$ epochs with a batch size of $6$ on PPI.
We do not increase the batch size after making decisions since PPI is small and only contains $20$ batches. The other parameters are the same as when searching on ModelNet10.
The search takes around $0.003$ day ($4$ minutes) on a Nvidia Tesla V100 GPU (16GB).

\savespace
\subsubsection{Architecture Evaluation on PPI}
We evaluate architectures on the PPI test set. We report the mean, standard derivation and the best accuracy and compare them with the SOTA methods and Random Search.
We also conduct an ablation study on number of cells and channel size which we include in the \supp.

\mysection{Training Settings}
We stack the discovered cell $5$ times with channel size 512.
Adam is used to optimize the model weights with initial learning rate $0.002$.
We use a cosine annealing to schedule the learning rate. 
Our architectures are trained for $2000$ epochs with batch size of $1$ as suggested in DeepGCNs \cite{Li2019DeepGCNsMG}. 
We find the best model on the validation dataset and obtain the micro-F1 score on the test dataset.

\begin{table}
\small
\centering
\resizebox{\columnwidth}{!}{%
\begin{tabular}{lccc}
\toprule
\textbf{Architecture} & \textbf{micro-F1} & \textbf{Params} & \textbf{Search Cost} \\ 
&                           \textbf{(\%)} & \textbf{(M)} & \textbf{(GPU-days)}\\
\midrule
GraphSAGE (LSTM) \cite{hamilton2017inductive} & 61.2 & 0.26 & manual\\ 
GeniePath \cite{liu2019geniepath} & 97.9 & 1.81 & manual \\ 
GAT \cite{velivckovic2017graph} & 97.3$\pm$0.2 & 3.64 & manual \\ 
DenseMRGCN-14 \cite{Li2019DeepGCNsMG} & 99.43 & 53.42 & manual \\ 
ResMRGCN-28 \cite{Li2019DeepGCNsMG} & 99.41 & 14.76 & manual \\  
\midrule
Random Search & 99.36$\pm$0.04 & 23.70 & random \\ 
SGAS (Cri.1 avg.) & 99.38$\pm$0.17 & 25.01 & 0.003 \\ 
SGAS (Cri.1 best) & \textbf{99.46} & 23.18 & 0.003 \\ 
SGAS (Cri.2 avg.) & 99.40$\pm$0.09 & 25.93 & 0.003 \\ 
SGAS (Cri.2 best) & \textbf{99.46} & 29.73 & 0.003 \\ 
SGAS (small) & 98.89 & 0.40 & 0.003 \\ 
\bottomrule
\end{tabular}
}
\vspace{3pt}
\caption{\textbf{Comparison with state-of-the-art architectures for node classification on PPI.} SGAS (small) is the small network stacking the cell searched by SGAS (Cri.1).}
\label{tab:ppi_comparsion_sota}
\savespace
\end{table}

\mysection{Evaluation Results and Analysis}
We compare the average and best performance of SGAS to other state-of-the-arts methods and Random Search on node classification on the PPI dataset.
\tblLabel \ref{tab:ppi_comparsion_sota} shows the best architecture discovered by our SGAS outperforms the state-of-the-art DenseMRGCN-14 \cite{Li2019DeepGCNsMG} by $\sim$0.03\% with $\sim$30.24 M less parameters. 
The average performance of SGAS also surpasses the Random Search baseline consistently. 
In addition, SGAS (Cri.2 avg.) outperforms SGAS (Cri.1 avg.) in terms of both mean and standard deviation. This indicates that \textit{Criterion 2} provides more stable results. We visualize the architecture with the best performance in \figLabel \ref{fig:visualizations_gcn} (b).

\section{Conclusion}
\label{sec:conclusion}
In this work, we propose the Sequential Greedy Architectural Search (SGAS) algorithm to design architectures automatically for CNNs and GCNs. The bi-level optimization problem in NAS is solved in a greedy fashion using heuristic criteria which take the edge importance, the selection certainty and the selection stability into consideration. Such an approach alleviates the effect of the \textbf{degenerate search-evaluation correlation} problem and reflects the true ranking of architectures. As a result, architectures discovered by SGAS achieve state-of-the-art performance on CIFAR-10, ImageNet, ModelNet and PPI datasets.

\mysection{Acknowledgments} This work was supported by the King Abdullah University of Science and Technology (KAUST) Office of Sponsored Research (OSR) through VCC funding. The authors  thank the KAUST IBEX team for helping to optimize the training workloads on ImageNet.

{\small
\bibliographystyle{ieee_fullname}
\bibliography{egbib}
}

\clearpage
\appendix

\twocolumn[{%
 \centering
 \LARGE SGAS: Sequential Greedy Architecture Search \\ -- Supplementary Material -- \\[1.5em]
}]

\section{Discussion} \label{appendix:supp}
\subsection{Greedy Methods in NAS}
The idea of incorporating greedy algorithms into NAS has been explored in several works. PNAS \cite{liu2018progressive} proposes a sequential model-based optimization (SMBO) approach to accelerate the search for CNN architectures. They start from a simple search space and a learn a predictor function. Then they greedily grow the search space by predicting scores of candidates cells using the learned predictor function. GNAS \cite{huang2018gnas} learns a global tree-like architecture for multi-attribute learning by iteratively updating layer-wise local architectures in a greedy manner. P-DARTS \cite{chen2019progressive} can also be regarded as a greedy approach, in which they bridge the depth gap between search and evaluation by gradually increasing the depth of the search networks while shrinking the number of candidate operations.

\subsection{Selection Criteria and Hyper-parameters}
\mysection{Edge Importance and Selection Certainty} Edge importance and selection certainty are combined into a single criterion, since the algorithm will be agnostic to the selection distribution of an edge, if we only consider edge importance. In this case, an edge may be selected with a sub-optimal operation at early epochs. On the other hand, we need to select $8$ out of $14$ edges in a DAG with $4$ intermediate nodes for a fair comparison with DARTS. Only considering selection certainty may fail to select the optimal $8$ edges, since an edge with a high selection certainty may have a high weight on \emph{Zero} operation (low edge importance).

\mysection{Choices of Hyper-parameters} Three extra parameters are introduced in \textit{SGAS}: \textbf{(1)} \emph{length of warm-up phase} \textbf{(2)} \emph{interval of greedy decisions} \textbf{(3)} \emph{history window size for Cri.2}. We provide a discussion on the default choices of them:

\noindent \textbf{(1)} Since the softmax weights of operations are initialized under a uniform distribution, choosing an operation for an edge after a short period of warming up leads to stable results. We simply set the \emph{length of the warm-up phase} to 9 epochs so that the first greedy decision will be made at the $10$th epoch. \textbf{(2)} For CNN experiments, the \emph{interval between greedy decisions} is chosen to be $5$. Since designing a normal cell with $4$ intermediate nodes needs to select $8$ out of $14$ edges ($8$ decisions to be made). For a fair comparison to our baseline DARTS, we want the search phase to last up to $50$ epochs, which is the length of search epochs in DARTS. For GCN architectures, in order to learn a compact network, we search a normal cell with $3$ intermediate nodes. Thus, we have $6$ decisions to make ($6$ out of $9$ edges). Similarly, to keep the length of the search phase less than 50 epochs, we set the \emph{interval between greedy decisions} to be $7$.
\textbf{(3)} The \emph{history window size for Cri.2} is always set as $4$, which is simply chosen to be slightly smaller than the \emph{interval between greedy decisions}.

\mysection{Ablation Study on Hyper-parameters}
In order to better understand the effects of the choices of hyper-parameters, we conduct ablation studies on \emph{interval of greedy decisions} $T$ and \emph{history window size $K$ for Cri.2} on CIFAR-$10$ in \tblLabel \ref{tab:cifar_ablation}. The default values of $T$ and $K$ are $5$ and $4$ respectively. We find that larger $T$ and $K$ stabilize the search and produce standard deviations in the test error. The test error only has a standard deviation as $0.08$ when $T=7$. When $T=3$, the average test error increases significantly from $2.67\%$ to $2.86\%$. We also find $K$ is less sensitive than $T$.

\begin{table}[h]
\small 
\setlength{\tabcolsep}{3pt}
\centering
\begin{tabular}{c|c|cc|cc}
\toprule
& & \multicolumn{2}{c|}{\textbf{Avg.}} & \multicolumn{2}{c}{\textbf{Best}} \\
$T$ & $K$ & \textbf{Params (M)} & \textbf{Test Err.(\%)} & \textbf{Params (M)} & \textbf{Test Err.(\%)} \\
\midrule
5 & 4 & 3.91$\pm$0.22 & 2.67$\pm$0.21  & 4.09 & 2.44\\
\midrule
3 & 4  & 4.09$\pm$0.24 & 2.86$\pm$0.12 & 4.39 & 2.69\\
7 & 4  & 3.66$\pm$0.16 & 2.65$\pm$0.08 & 3.68 & 2.54\\
\midrule
5 & 2  & 3.87$\pm$0.20 & 2.73$\pm$0.16 & 3.94 & 2.51\\
5 & 6  & 3.93$\pm$0.26 & 2.67$\pm$0.17 & 3.70 & 2.47\\
\bottomrule
\end{tabular}
\vspace{3pt}
\caption{\textbf{Ablation study on \emph{interval of greedy decisions} $T$ and \emph{history window size $K$ for Cri.2} on CIFAR-$10$.} We use SGAS (Cri.2) as our search method. We report the average and best performance of searched architectures.}
\label{tab:cifar_ablation}
\end{table}

\section{Experimental Details}
\subsection{GCN Experiments}
\mysection{GCN operators}
Similar as the search for CNN, SGAS selects one operation from a candidate operation search space for each edge in the DAG. We choose the following $10$ operations as our candidate operations: \textit{conv-1$\times$1}, \textit{MRConv} \cite{Li2019DeepGCNs}, \textit{EdgeConv} \cite{dgcnn}, \textit{GAT} \cite{velivckovic2017graph}, \textit{SemiGCN} \cite{Kipf2016SemiSupervisedCW}, \textit{GIN} \cite{Xu2018HowPAGIN}, \textit{SAGE} \cite{hamilton2017inductive}, \textit{RelSAGE}, \textit{skip-connect}, and \textit{zero} operation. \textit{conv-1$\times$1} is a basic convolution operation without aggregating information from neighbors, which is similar to PointNet \cite{Qi2016PointNetDL}. \textit{MRConv} \cite{Li2019DeepGCNs}, \textit{EdgeConv} \cite{dgcnn}, \textit{GAT} \cite{velivckovic2017graph}, \textit{SemiGCN} \cite{Kipf2016SemiSupervisedCW}, \textit{GIN} \cite{Xu2018HowPAGIN} and \textit{SAGE} \cite{hamilton2017inductive} are widely used GCN operators in the graph learning domain and the 3D computer vision domain. RelSAGE is a modified GraphSAGE (SAGE) \cite{hamilton2017inductive} which combines the ideas from \textit{MRConv} \cite{Li2019DeepGCNs} and GraphSAGE \cite{hamilton2017inductive}. Instead of aggregating the node features with its neighbor features directly, we aggregate the node features with the difference between the node features and its neighbor features:

\resizebox{0.91\hsize}{!}{%
$\mathbf{h}_{v}^{(k)}=\sigma\left(\mathbf{W}^{(k)} \cdot f_{k}\left(\mathbf{h}_{v}^{(k-1)},\left\{\mathbf{h}_{u}^{(k-1)} - \mathbf{h}_{v}^{(k-1)}, \forall u \in \mathcal{N}(v)\right\}\right)\right)$}
where $\mathbf{h}_{v}^{(k)}$ is the feature of the center node $v$ in $k$-th layer. $\mathcal{N}(v)$ denotes the neighbors of node $v$. $f_k$ is a max aggregation function and $\sigma$ is a ReLU activation function as GraphSAGE \cite{hamilton2017inductive}. The GCNs operators are implemented using Pytorch Geometric \cite{fey2019fast}. We also add \textit{skip-connect} (similar as residual graph connections \cite{Li2019DeepGCNs}) and \textit{zero} operation in our search space.

\mysection{Ablation Study on GCN Cells}
We conduct an ablation study on the parameter size of the best cell searched on PPI by SGAS (Cri.1 best). 
\tblLabel \ref{tab:ppi_ablation} shows the trade-off between the parameter size and the final performance. To derive a compact model, we can use a smaller number of cells or less channels in the architecture searched by SGAS.

\begin{table}[h]
\small 
\setlength{\tabcolsep}{3pt}
\centering
\begin{tabular}{cc|cc}
\toprule
\textbf{Number of Cells} &\textbf{Channel Size} & \textbf{Params (M)} & \textbf{micro-F1(\%)} \\
\midrule
5 & 64   & 0.40 & 98.894  \\
5 & 128  & 1.52 & 99.369  \\
5 & 256  & 5.89 & 99.429\\
5 & 512  & 23.18 & 99.462 \\
\midrule
1   & 256   & 1.22  & 99.157 \\
3   & 256   & 3.52  & 99.418  \\
5   & 256   & 5.89  & 99.429\\
7   & 256   & 8.25  & 99.433 \\
\bottomrule
\end{tabular}
\vspace{3pt}
\caption{\textbf{Ablation study on channel size and number of cells on node classification on PPI.} We use SGAS (Cri.1 best), the best architecture we discovered by using \textit{Criterion 1} to conduct experiments.}
\label{tab:ppi_ablation}
\end{table}

\subsection{More Details}
\begin{figure*}[h]
 \centering
 \begin{tabular}{@{}c@{\hspace{1mm}}c@{\hspace{1mm}}c@{}}
 
        \includegraphics[width=\columnwidth]{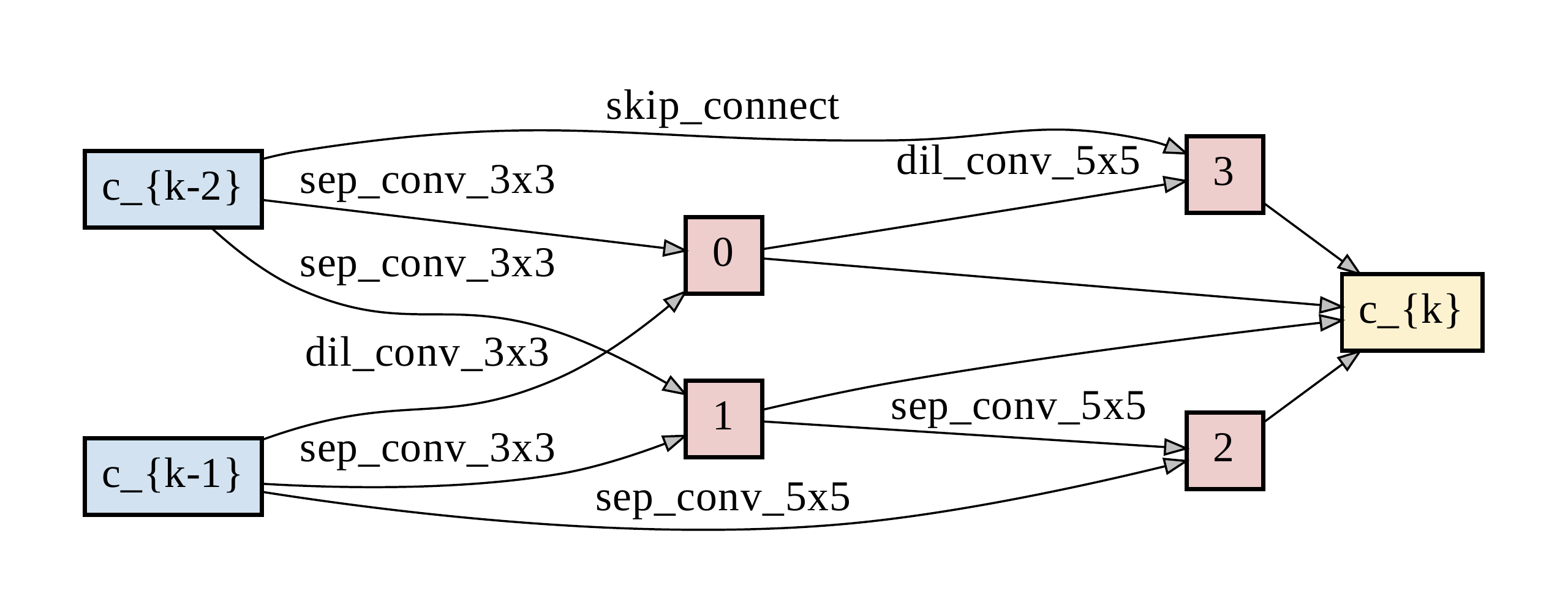} &
 		\includegraphics[width=\columnwidth]{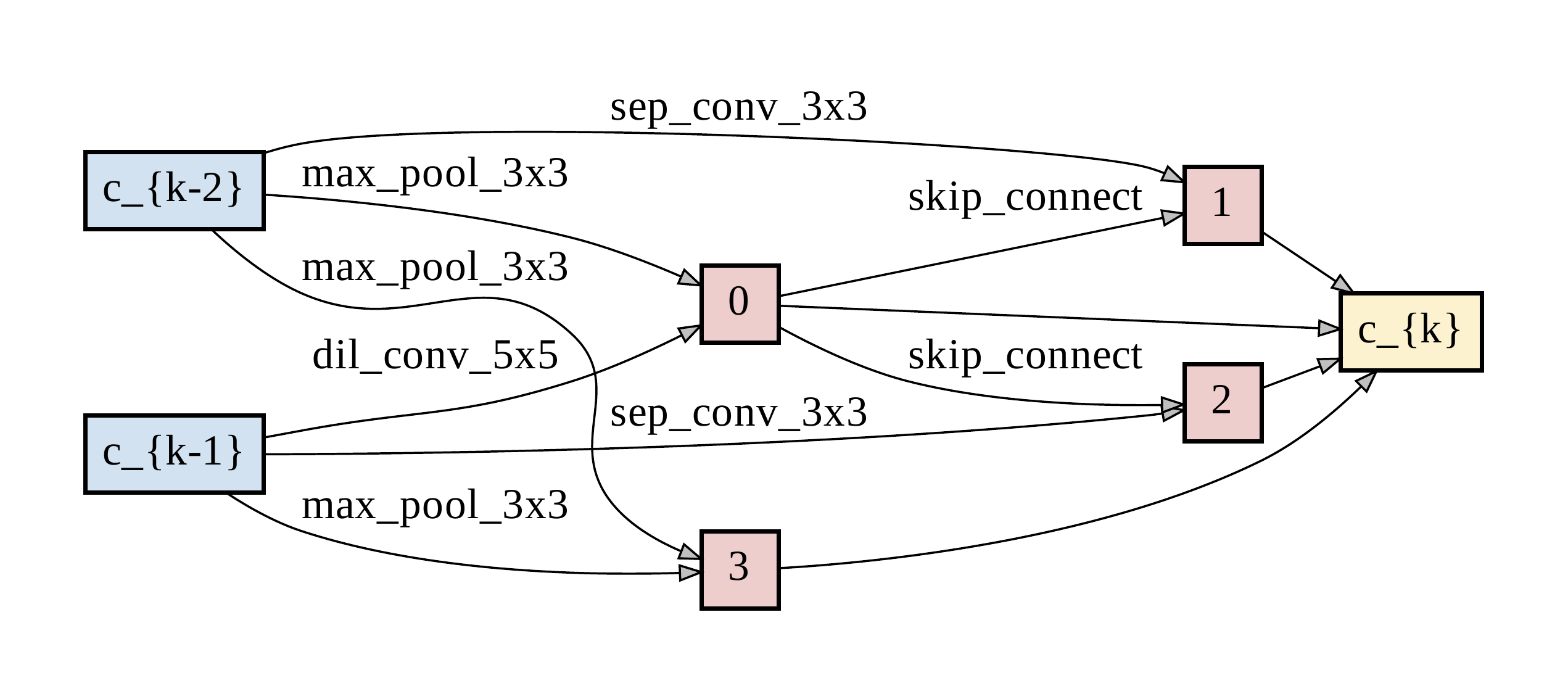} \\
		
 		\small (a) Normal cell of the best model with SGAS \textit{Cri. 1} &
 		\small (b) Reduction cell of the best model with SGAS \textit{Cri. 1}\\
 		\small  on CIFAR-10 and ImageNet &
 		\small on CIFAR-10 and ImageNet\\
 		
 		\includegraphics[width=\columnwidth]{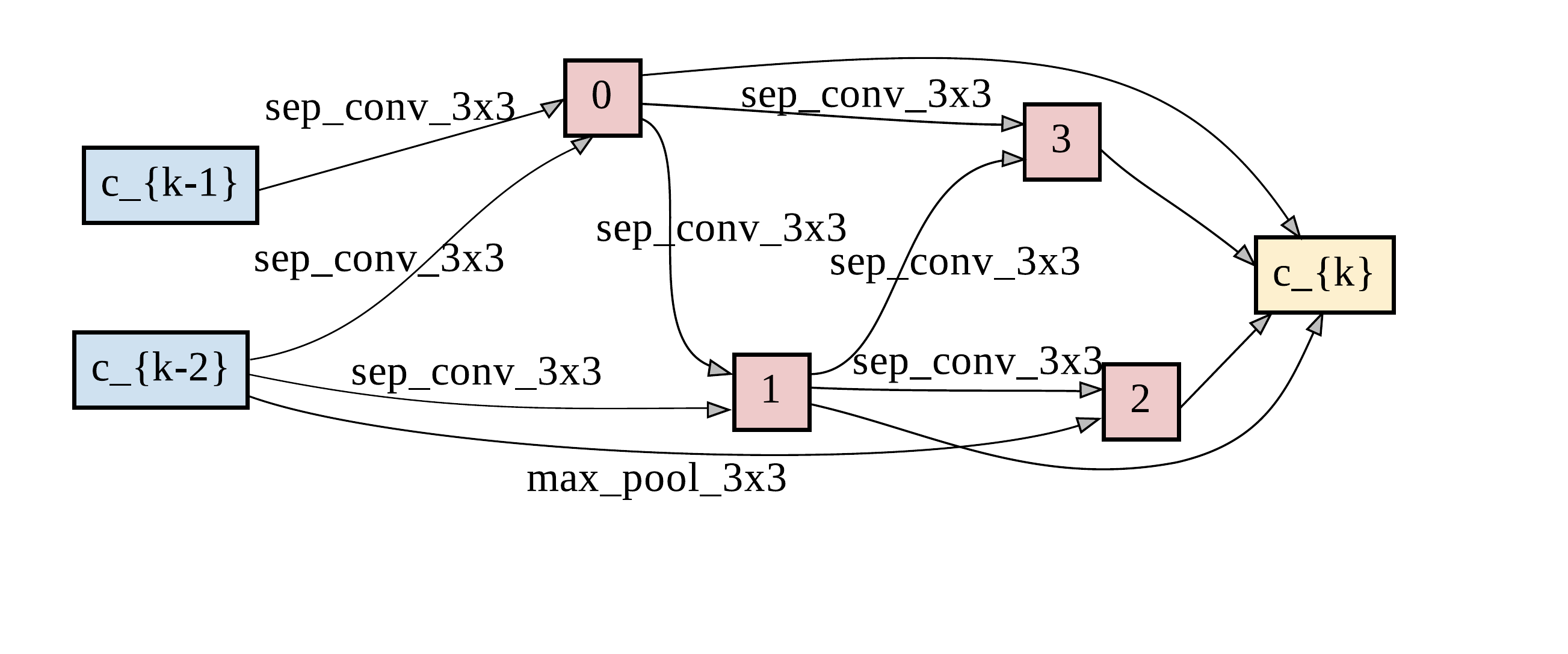} &
 		\includegraphics[height=0.45\columnwidth]{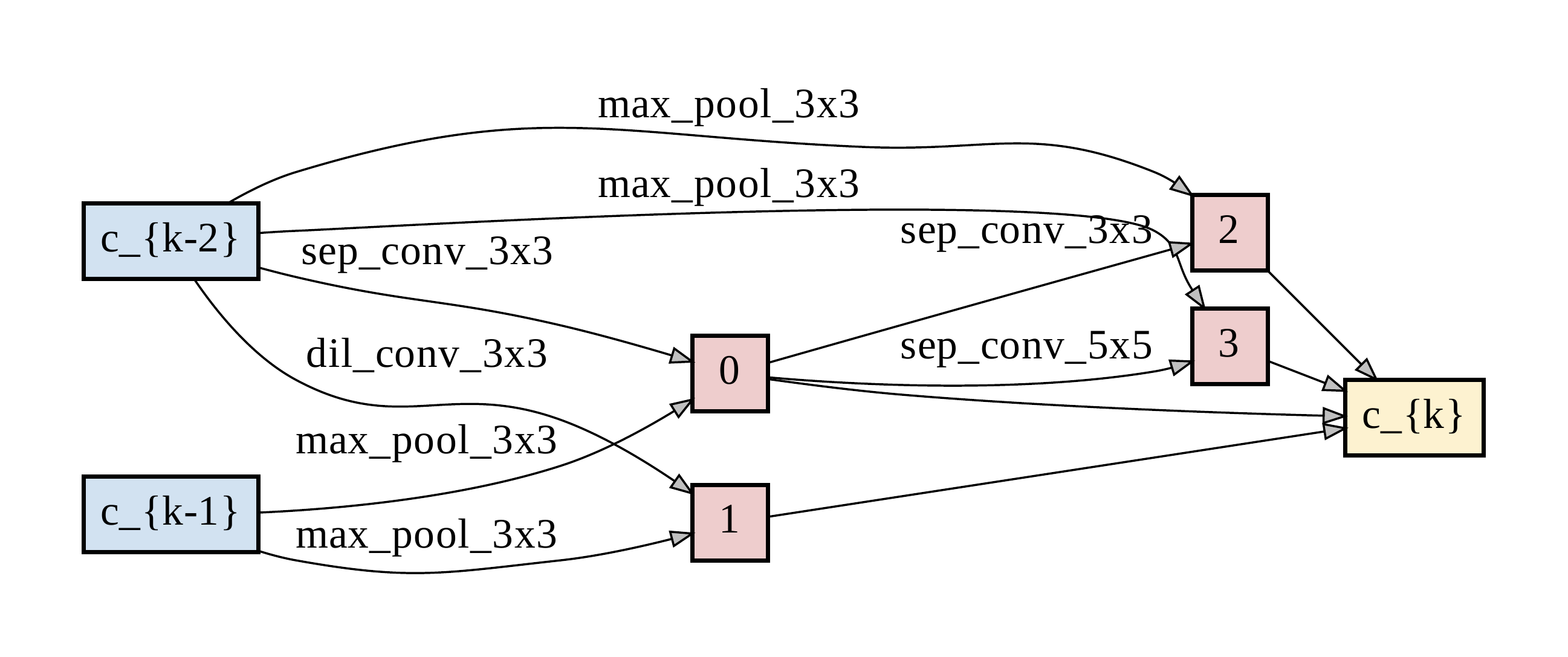} \\
		
 		\small (c) Normal cell of the best model with SGAS \textit{Cri. 2} on CIFAR-10&
 		\small (d) Reduction cell of the best model with SGAS \textit{Cri. 2} on CIFAR-10\\
 		 
 		\includegraphics[width=\columnwidth]{figures/visualizations/hist_imagenet_normal.pdf} &
 		\includegraphics[height=0.45\columnwidth]{figures/visualizations/hist_imagenet_reduction.pdf} \\
		
 		\small (e) Normal cell of the best model with SGAS \textit{Cri. 2} on ImageNet &
 		\small (f) Reduction cell of the best model with SGAS \textit{Cri. 2} on ImageNet\\

 \end{tabular}
 \caption{\textbf{Best cell architecture for image classification tasks}}
 \label{fig:visualizations_cnn_all}
 \end{figure*}

\begin{figure*}
 \centering
 \begin{tabular}{@{}c@{\hspace{1mm}}c@{\hspace{1mm}}c@{}}
 		
 		\includegraphics[width=\columnwidth]{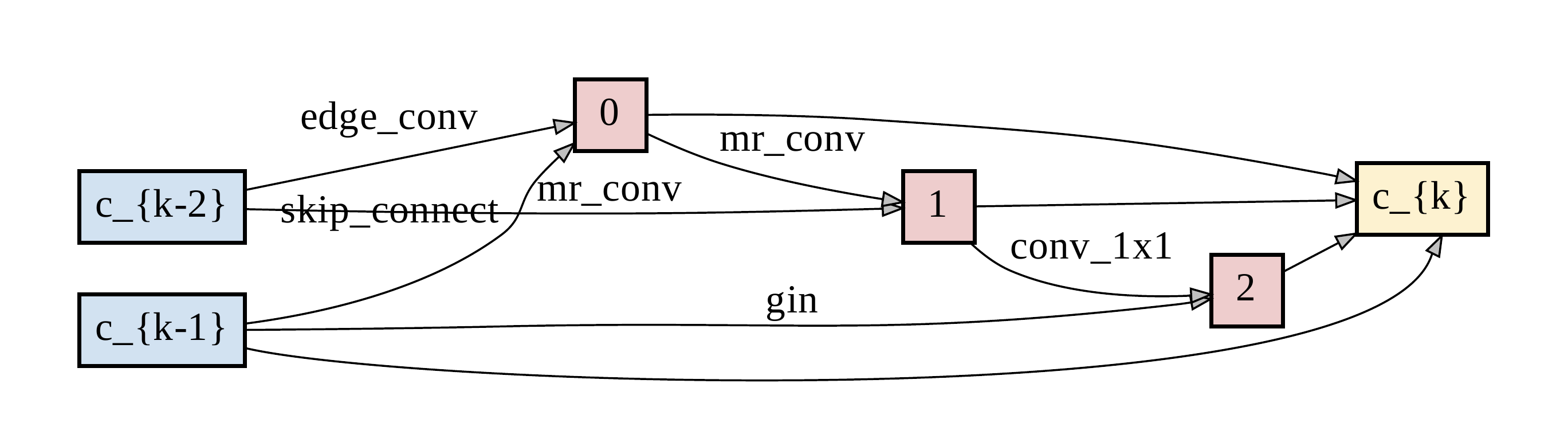} &
 		\includegraphics[width=\columnwidth]{figures/visualizations/hist_modelnet_normal.pdf}\\
 		
 		\small (a) Normal cell of the best model with SGAS \textit{Cri. 1} on ModelNet40 &
		\small (b) Normal cell of the best model with SGAS \textit{Cri. 2} on ModelNet40\\

 		\includegraphics[width=\columnwidth]{figures/visualizations/nohis_ppi_normal.pdf} &
        \includegraphics[width=\columnwidth]{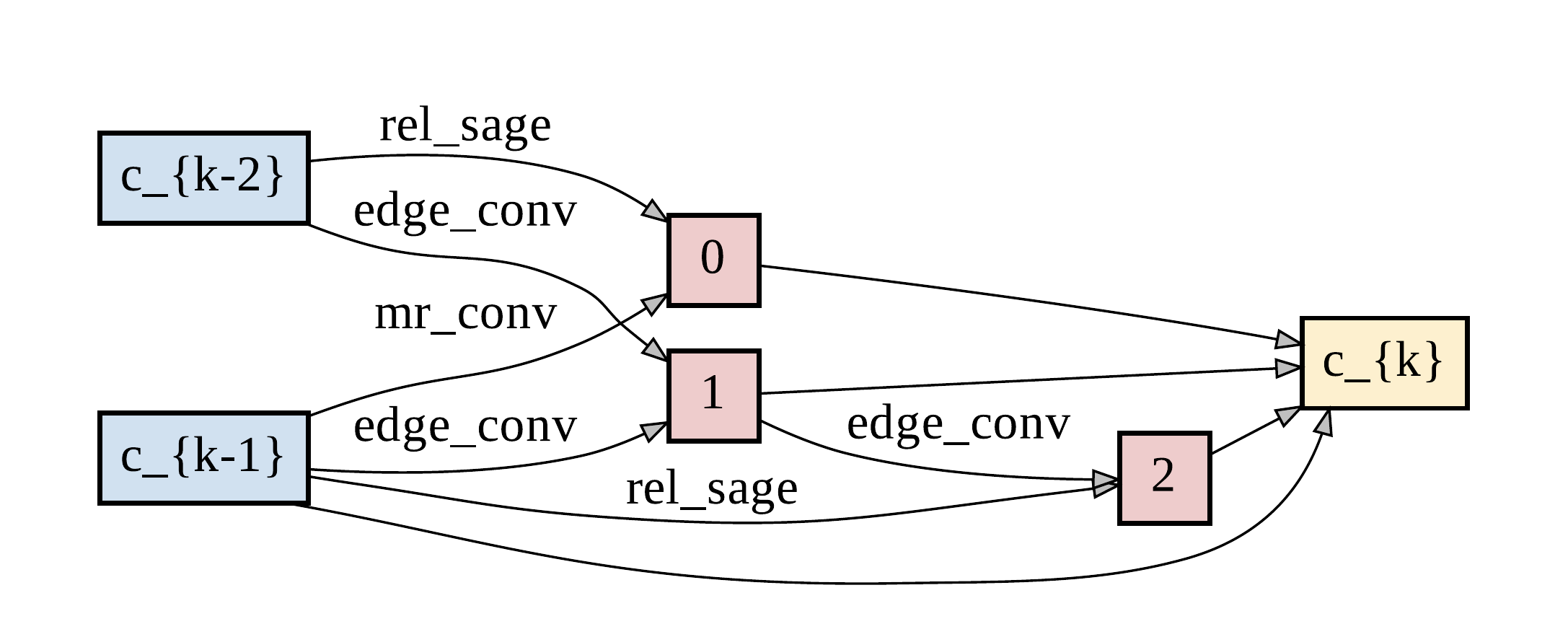}\\
		
		\small (c) Normal cell of the best model with SGAS \textit{Cri. 1} on PPI &
		\small (d) Normal cell of the best model with SGAS \textit{Cri. 2} on PPI\\

 \end{tabular}
 \caption{\textbf{Best cell architectures on ModelNet40 and PPI with each \textit{Criterion}}}
 \savespace
 \label{fig:visualizations_gcn_all}
 \end{figure*}

\mysection{Cell Visualizations} We visualize the best cells discovered by SGAS with different criteria (\textit{Criterion 1} and \textit{Criterion 2}) mentioned in the experiment section. \figLabel  \ref{fig:visualizations_cnn_all} shows the best cells for CNNs on CIFAR-$10$ and ImageNet. \figLabel \ref{fig:visualizations_gcn_all} shows the best cells for GCNs on ModelNet-$40$ and PPI.

\mysection{Detailed results}
Here we provide the detailed results mentioned in the experimental section of the paper. In the CNN experiments, we compare SGAS with DARTS on CIFAR-$10$ and ImageNet. We execute the search phase $10$ times for both SGAS (Cri.1 and Cri.2) and DARTS (1st and 2nd order) to obtain $10$ different architectures per method. For each resulting architecture, we run the evaluation phase and assign a ranking based on the evaluation accuracy. To measure the discrepancy between the search and evaluation, we calculate the Kendall $\tau$ correlation between the ranking of the search phase and the evaluation phase.
We show these results in \tblLabel \ref{tbl:detail_crit1_cifar10} and \tblLabel \ref{tbl:detail_crit2_cifar10} for SGAS, and \tblLabel \ref{tbl:detail_darts1st_cifar10} and \tblLabel \ref{tbl:detail_darts2nd_cifar10} for DARTS.
For ImageNet, we evaluate the top three architectures found on CIFAR-$10$. We show the results in \tblLabel \ref{tbl:detail_crit1_imagenet} and \tblLabel \ref{tbl:detail_crit2_imagenet} for both \textit{Criterion 1} and \textit{Criterion 2}.

In the GCN experiments, we compare SGAS (Cri.1 and Cri.2) with a random search baseline on ModelNet and PPI. Similar as in experiments for CNNs, we conduct the search phase $10$ times for each method. For experiments on ModelNet, we search cells on ModelNet10 and then evaluate the searched cells on ModelNet40. The results are shown for \textit{Criterion 1}, \textit{Criterion 2} and random search in \tblLabel \ref{tbl:detail_crit1_modelnet}, \tblLabel \ref{tbl:detail_crit2_modelnet} and \tblLabel \ref{tbl:detail_random_modelnet} respectively. The results on PPI are presented in \tblLabel \ref{tbl:detail_crit1_ppi} and \tblLabel \ref{tbl:detail_crit2_ppi} for each \textit{Criterion} and in \tblLabel \ref{tbl:detail_random_ppi} for random search.

\begin{table*}
\centering
\begin{tabular}{lccc|c}

\toprule
\textbf{Experiment} & \textbf{Validation error (\%)} & \textbf{Params (M)} & \textbf{Test error (\%)} & \textbf{Evaluation ranking} \\ 
\midrule
Cri.1\_CIFAR\_1 & 16.94 & 3.75 & 2.44 & 2 \\ 
Cri.1\_CIFAR\_2 & 17.33 & 3.73 & 2.50 & 3 \\ 
Cri.1\_CIFAR\_3 & 17.90 & 3.80 & 2.39 & 1 \\ 
Cri.1\_CIFAR\_4 & 17.90 & 3.32 & 2.63 & 6 \\ 
Cri.1\_CIFAR\_5 & 17.99 & 3.45 & 2.78 & 8 \\ 
Cri.1\_CIFAR\_6 & 18.43 & 3.47 & 2.68 & 7 \\ 
Cri.1\_CIFAR\_7 & 18.72 & 3.83 & 2.51 & 4 \\ 
Cri.1\_CIFAR\_8 & 19.82 & 3.66 & 2.61 & 5 \\ 
Cri.1\_CIFAR\_9 & 19.93 & 3.98 & 3.18 & 10 \\ 
Cri.1\_CIFAR\_10 & 21.53 & 3.61 & 2.87 & 9 \\ 
\midrule
\textbf{Average} & 18.65$\pm$1.4 & 3.66$\pm$0.2 & 2.66$\pm$0.24 & \textbf{Kendall $\tau$}\\\cmidrule{5-5}
\textbf{Best Model} & 17.90 & 3.80 & 2.39 & 0.56 \\ 
\bottomrule
\end{tabular}
\vspace{3pt}
\caption{\textbf{Results of SGAS \textit{Criterion 1} on CIFAR-10}}
\label{tbl:detail_crit1_cifar10}
\end{table*}

\begin{table*}
\centering
\begin{tabular}{lccc|c}

\toprule
\textbf{Experiment} & \textbf{Validation error (\%)} & \textbf{Params (M)} & \textbf{Test error (\%)} & \textbf{Evaluation ranking} \\ 
\midrule
Cri.2\_CIFAR\_1 & 16.48 & 4.14 & 2.57 & 4 \\ 
Cri.2\_CIFAR\_2 & 17.26 & 3.88 & 2.60 & 6 \\ 
Cri.2\_CIFAR\_3 & 17.31 & 4.09 & 2.44 & 1 \\ 
Cri.2\_CIFAR\_4 & 17.47 & 3.91 & 2.49 & 2 \\ 
Cri.2\_CIFAR\_5 & 17.53 & 3.69 & 2.52 & 3 \\ 
Cri.2\_CIFAR\_6 & 17.98 & 3.95 & 3.12 & 10 \\ 
Cri.2\_CIFAR\_7 & 18.28 & 3.69 & 2.58 & 5 \\ 
Cri.2\_CIFAR\_8 & 18.28 & 4.33 & 2.85 & 8 \\ 
Cri.2\_CIFAR\_9 & 19.48 & 3.73 & 2.85 & 9 \\ 
Cri.2\_CIFAR\_10 & 19.98 & 3.68 & 2.66 & 7 \\ 
\midrule
\textbf{Average} & 18.00$\pm$1.06 & 3.91$\pm$0.22 & 2.67$\pm$0.21 & \textbf{Kendall $\tau$}\\\cmidrule{5-5}
\textbf{Best Model} & 17.31 & 4.09 & 2.44 & 0.42 \\
\bottomrule
\end{tabular}
\vspace{3pt}
\caption{\textbf{Results of SGAS \textit{Criterion 2} on CIFAR-10}}
\label{tbl:detail_crit2_cifar10}
\end{table*}

\begin{table*}
\centering
\begin{tabular}{lccc|c}

\toprule
\textbf{Experiment} & \textbf{Validation error (\%)} & \textbf{Params (M)} & \textbf{Test error (\%)} & \textbf{Evaluation ranking} \\ 
\midrule
DARTS\_1st\_CIFAR\_1 & 11.37 & 3.27 & 2.83 & 4 \\
DARTS\_1st\_CIFAR\_2 & 11.45 & 3.65 & 2.57 & 2 \\
DARTS\_1st\_CIFAR\_3 & 11.47 & 2.29 & 2.94 & 7 \\
DARTS\_1st\_CIFAR\_4 & 11.48 & 2.65 & 2.96 & 8 \\
DARTS\_1st\_CIFAR\_5 & 11.65 & 3.09 & 2.50 & 1 \\
DARTS\_1st\_CIFAR\_6 & 11.75 & 2.86 & 2.84 & 5 \\
DARTS\_1st\_CIFAR\_7 & 11.77 & 2.09 & 3.06 & 10 \\
DARTS\_1st\_CIFAR\_8 & 11.81 & 2.52 & 3.01 & 9 \\
DARTS\_1st\_CIFAR\_9 & 11.82 & 2.65 & 2.94 & 6 \\
DARTS\_1st\_CIFAR\_10 & 11.94 & 3.27 & 2.82 & 3 \\
\midrule
\textbf{Average} & 11.65$\pm$0.19 & 2.84$\pm$0.49 & 2.85$\pm$0.18 & \textbf{Kendall $\tau$} \\\cmidrule{5-5}
\textbf{Best Model} & 11.65 & 3.09 & 2.50 & 0.16 \\
\bottomrule
\end{tabular}
\vspace{3pt}
\caption{\textbf{Results of DARTS 1st order on CIFAR-10}}
\label{tbl:detail_darts1st_cifar10}
\end{table*}

\begin{table*}
\centering
\begin{tabular}{lccc|c}

\toprule
\textbf{Experiment} & \textbf{Validation error (\%)} & \textbf{Params (M)} & \textbf{Test error (\%)} & \textbf{Evaluation ranking} \\ 
\midrule
DARTS\_2nd\_CIFAR\_1 & 11.35 & 2.91 & 2.96 & 8 \\
DARTS\_2nd\_CIFAR\_2 & 11.51 & 2.93 & 2.73 & 5 \\
DARTS\_2nd\_CIFAR\_3 & 11.68 & 2.20 & 3.01 & 9 \\
DARTS\_2nd\_CIFAR\_4 & 11.76 & 2.66 & 2.75 & 6 \\
DARTS\_2nd\_CIFAR\_5 & 11.80 & 3.09 & 2.72 & 4 \\
DARTS\_2nd\_CIFAR\_6 & 11.82 & 3.40 & 2.62 & 3 \\
DARTS\_2nd\_CIFAR\_7 & 11.83 & 2.91 & 2.82 & 7 \\
DARTS\_2nd\_CIFAR\_8 & 11.93 & 3.20 & 2.51 & 1 \\
DARTS\_2nd\_CIFAR\_9 & 11.95 & 2.14 & 3.48 & 10 \\
DARTS\_2nd\_CIFAR\_10 & 12.03 & 2.55 & 2.62 & 2 \\
\midrule
\textbf{Average} & 11.77$\pm$0.21 & 2.8$\pm$0.41 & 2.82$\pm$0.28 & \textbf{Kendall $\tau$} \\\cmidrule{5-5}
\textbf{Best Model} & 11.93 & 3.20 & 2.51 & -0.29\\
\bottomrule
\end{tabular}
\vspace{3pt}
\caption{\textbf{Results of DARTS 2nd order on CIFAR-10}}
\label{tbl:detail_darts2nd_cifar10}
\end{table*}

\begin{table*}
\centering
\begin{tabular}{lcccc}
\toprule

\textbf{Experiment} & \textbf{Test error top-1 (\%)} & \textbf{Test error top-5 (\%)} & \textbf{Params (M)} & \textbf{$\times+$} \\
\midrule
Cri.1\_ImageNet\_1 & 24.47 & 7.23 & 5.25 & 578 \\
Cri.1\_ImageNet\_2 & 24.53 & 7.40 & 5.23 & 574 \\
Cri.1\_ImageNet\_3 & 24.22 & 7.25 & 5.29 & 585 \\
\midrule
\textbf{Average} & 24.41$\pm$0.16 & 7.29$\pm$0.09 & 5.25$\pm$0.03 & 579 \\
\textbf{Best Model} & 24.22 & 7.25 & 5.29 & 585 \\
\bottomrule
\end{tabular}
\vspace{3pt}
\caption{\textbf{Results of SGAS with \textit{Criterion 1} on ImageNet.} Note that the chosen architectures are the three best ones from the results obtained on CIFAR-10.}
\label{tbl:detail_crit1_imagenet}
\end{table*}

\begin{table*}
\centering
\begin{tabular}{lcccc}
\toprule

\textbf{Experiment} & \textbf{Test error top-1 (\%)} & \textbf{Test error top-5 (\%)} & \textbf{Params (M)} & \textbf{$\times+$} \\
\midrule
Cri.2\_ImageNet\_1 & 24.44 & 7.41 & 5.70 & 621 \\
Cri.2\_ImageNet\_2 & 24.13 & 7.31 & 5.44 & 598 \\
Cri.2\_ImageNet\_3 & 24.55 & 7.44 & 5.20 & 571 \\
\midrule
\textbf{Average} & 24.38$\pm$0.22 & 7.39$\pm$0.07 & 5.44$\pm$0.25 & 597 \\
\textbf{Best Model} & 24.13 & 7.31 & 5.44 & 598 \\

\bottomrule
\end{tabular}
\vspace{3pt}
\caption{\textbf{Results of SGAS with \textit{Criterion 2} on ImageNet.} Note that the chosen the architectures are the three best ones from the results obtained on CIFAR-10.}
\label{tbl:detail_crit2_imagenet}
\end{table*}

\begin{table*}
\centering
\begin{tabular}{lcc}
\toprule

\textbf{Experiment} & \textbf{Params (M)} & \textbf{Test OA (\%)} \\
\midrule
Cri.1\_ModelNet\_1 & 8.79 &	92.71 \\
Cri.1\_ModelNet\_2 & 9.23 &	92.83 \\
Cri.1\_ModelNet\_3 & 8.79 &	92.79 \\
Cri.1\_ModelNet\_4 & 8.78 &	92.34 \\
Cri.1\_ModelNet\_5 & 8.93 &	92.79 \\
Cri.1\_ModelNet\_6 & 8.19 &	92.30 \\
Cri.1\_ModelNet\_7 & 8.63 &	92.83 \\
Cri.1\_ModelNet\_8 & 8.63 &	92.71 \\
Cri.1\_ModelNet\_9 & 8.63 &	92.87 \\
Cri.1\_ModelNet\_10 & 9.23 & 92.79 \\
\midrule
\textbf{Average} & 8.78$\pm$0.30 & 92.69$\pm$0.20 \\
\textbf{Best Model} & 8.63 & 92.87\\
\bottomrule
\end{tabular}
\vspace{3pt}
\caption{\textbf{Results of SGAS with \textit{Criterion 1} on ModelNet40.} Architectures are formed by stacking 9 cells with 128 channel size.}
\label{tbl:detail_crit1_modelnet}
\end{table*}

\begin{table*}
\centering
\begin{tabular}{lcc}
\toprule

\textbf{Experiment} & \textbf{Params (M)} & \textbf{Test OA (\%)}\\
\midrule
Cri.2\_ModelNet\_1 & 8.78 &	92.91 \\
Cri.2\_ModelNet\_2 & 8.78 &	92.67 \\
Cri.2\_ModelNet\_3 & 9.08 &	92.79 \\
Cri.2\_ModelNet\_4 & 8.49 &	93.23 \\
Cri.2\_ModelNet\_5 & 9.08 &	93.03 \\
Cri.2\_ModelNet\_6 & 9.08 &	93.07 \\
Cri.2\_ModelNet\_7 & 8.78 &	93.11 \\
Cri.2\_ModelNet\_8 & 8.63 &	92.67 \\
Cri.2\_ModelNet\_9 & 8.78 &	92.83 \\
Cri.2\_ModelNet\_10 & 9.23 & 92.95 \\
\midrule
\textbf{Average} & 8.87$\pm$0.23 & 92.93$\pm$0.19 \\
\textbf{Best Model} & 8.49 & 93.23 \\
\bottomrule
\end{tabular}
\vspace{3pt}
\caption{\textbf{Results of SGAS with \textit{Criterion 2} on ModelNet40.} Architectures are formed by stacking 9 cells with 128 channel size.}
\label{tbl:detail_crit2_modelnet}
\end{table*}

\begin{table*}
\centering
\begin{tabular}{lcc}
\toprule

\textbf{Experiment} & \textbf{Params (M)} & \textbf{Test OA (\%)}\\
\midrule
Random\_ModelNet\_1 & 9.22	&	92.79 \\
Random\_ModelNet\_2 & 8.93	&	92.67 \\
Random\_ModelNet\_3 & 9.08	&	92.71 \\
Random\_ModelNet\_4 & 8.78	&	92.46 \\
Random\_ModelNet\_5 & 8.19	&	92.79 \\
Random\_ModelNet\_6 & 8.63	&	92.54 \\
Random\_ModelNet\_7 & 8.93	&	91.94 \\
Random\_ModelNet\_8 & 8.63	&	92.99 \\
Random\_ModelNet\_9 & 8.79	&	93.15 \\
Random\_ModelNet\_10 & 8.49	&	92.46 \\
\midrule
\textbf{Average} & 8.77$\pm$0.30 & 92.65$\pm$0.33 \\
\textbf{Best Model} & 8.79 & 93.15\\
\bottomrule
\end{tabular}
\vspace{3pt}
\caption{\textbf{Results of random search on ModelNet40.} Architectures are formed by stacking 9 cells with 128 channel size.}
\label{tbl:detail_random_modelnet}
\end{table*}

\begin{table*}
\centering
\begin{tabular}{lccc|c}
\toprule

\textbf{Experiment} & \textbf{Params (M)} & \textbf{Test micro-F1 (\%)} \\
\midrule
Cri.1\_PPI\_1 & 27.11 & 99.45 \\
Cri.1\_PPI\_2 & 23.18 & 99.42 \\
Cri.1\_PPI\_3 & 25.80 & 98.91 \\
Cri.1\_PPI\_4 & 25.80 & 99.38 \\
Cri.1\_PPI\_5 & 24.49 & 99.44 \\
Cri.1\_PPI\_6 & 29.73 & 99.44 \\
Cri.1\_PPI\_7 & 24.50 & 99.44 \\
Cri.1\_PPI\_8 & 21.87 & 99.43 \\
Cri.1\_PPI\_9 & 24.49 & 99.44 \\
Cri.1\_PPI\_10 & 23.18 & 99.46 \\
\midrule
\textbf{Average} & 25.01$\pm$2.24 & 99.38$\pm$0.17 \\
\textbf{Best Model} & 23.18 & 99.46 \\
\bottomrule
\end{tabular}
\vspace{3pt}
\caption{\textbf{Results of SGAS with \textit{Criterion 1} on PPI.}}
\label{tbl:detail_crit1_ppi}
\end{table*}

\begin{table*}
\centering
\begin{tabular}{lcc}
\toprule

\textbf{Experiment} & \textbf{Params (M)} & \textbf{Test micro-F1 (\%)} \\
\midrule
Cri.2\_PPI\_1 & 25.80 & 99.17 \\
Cri.2\_PPI\_2 & 28.42 & 99.46 \\
Cri.2\_PPI\_3 & 20.55 & 99.40 \\
Cri.2\_PPI\_4 & 21.87 & 99.43 \\
Cri.2\_PPI\_5 & 24.49 & 99.44 \\
Cri.2\_PPI\_6 & 28.42 & 99.43 \\
Cri.2\_PPI\_7 & 29.73 & 99.42 \\
Cri.2\_PPI\_8 & 25.80 & 99.45 \\
Cri.2\_PPI\_9 & 28.42 & 99.44 \\
Cri.2\_PPI\_10 & 25.79 & 99.41 \\
\midrule
\textbf{Average} & 25.93$\pm$2.99 & 99.40$\pm$0.09\\
\textbf{Best Model} & 28.42 & 99.46 \\
\bottomrule
\end{tabular}
\vspace{3pt}
\caption{\textbf{Results of SGAS with \textit{Criterion 2} on PPI.}}
\label{tbl:detail_crit2_ppi}
\end{table*}

\begin{table*}
\centering
\begin{tabular}{lcc}
\toprule

\textbf{Experiment} & \textbf{Params (M)} & \textbf{Test micro-F1 (\%)} \\
\midrule
Random\_PPI\_1 & 20.57 & 99.27 \\
Random\_PPI\_2 & 24.48 & 99.37 \\
Random\_PPI\_3 & 24.49 & 99.40 \\
Random\_PPI\_4 & 19.24 & 99.40 \\
Random\_PPI\_5 & 24.48 & 99.37 \\
Random\_PPI\_6 & 21.85 & 99.36 \\
Random\_PPI\_7 & 27.11 & 99.32 \\
Random\_PPI\_8 & 23.17 & 99.40 \\
Random\_PPI\_9 & 24.48 & 99.39 \\
Random\_PPI\_10 & 27.11 & 99.38 \\
\midrule
\textbf{Average} & 23.7$\pm$2.56 & 99.36$\pm$0.04\\
\textbf{Best Model} & 23.17 & 99.40\\
\bottomrule
\end{tabular}
\vspace{3pt}
\caption{\textbf{Results of random search on PPI.}}
\label{tbl:detail_random_ppi}
\end{table*}
\end{document}